%% file: main.tex
\documentclass[10pt,twocolumn,letterpaper]{article}

\PassOptionsToPackage{table}{xcolor}
\usepackage[final]{cvpr}

\input{preamble}

\usepackage[utf8]{inputenc}
\usepackage[T1]{fontenc}
\usepackage{url}
\usepackage{amsfonts}
\usepackage{nicefrac}
\usepackage{microtype}
\usepackage{tabularx}
\usepackage{multirow}
\usepackage{soul}
\usepackage{multirow}
\usepackage[normalem]{ulem}
\usepackage{enumitem}
\useunder{\uline}{\ul}{}
\usepackage{subcaption}
\usepackage{pifont}
\newcommand{\cmark}{\ding{51}}
\newcommand{\xmark}{\ding{55}}

\definecolor{promptbg}{HTML}{F8F9FA}
\definecolor{promptframe}{HTML}{CED4DA}
\definecolor{systemtitle}{HTML}{495057}
\definecolor{evaltitle}{HTML}{4A4E69}

\input{prompt_box}

\AtEndPreamble{%
\Crefname{figure}{Fig.}{Figs.}
\Crefname{table}{Tab.}{Tabs.}
\Crefname{section}{Sec.}{Secs.}
\Crefname{subsection}{Sec.}{Secs.}
\Crefname{equation}{Eq.}{Eqs.}
}

\definecolor{mask_orange}{RGB}{239, 134, 50}

\definecolor{mask_purple}{RGB}{169, 80, 246}

\definecolor{mask_yellow}{RGB}{255, 255, 97}

\definecolor{mask_blue}{RGB}{55, 125, 247}

\definecolor{mask_red}{RGB}{201, 57, 31}

\makeatletter
\DeclareRobustCommand\onedot{\futurelet\@let@token\@onedot}
\def\@onedot{\ifx\@let@token.\else.\null\fi\xspace}

\def\eg{\emph{e.g}\onedot} 
\def\ie{\emph{i.e}\onedot}

\def\etal{\emph{et al}\onedot}
\makeatother

\definecolor{revorange}{rgb}{0.94,0.56,0.06}
\newif\ifshowrev\showrevtrue

\definecolor{revblue}{rgb}{0.28,0.35,0.93}
\newif\ifshowrevok\showrevoktrue

\newcommand{\blname}[1]{\textcolor{gray}{#1}}

\definecolor{codepurple}{rgb}{0.58,0,0.82}
\definecolor{highlight_green}{RGB}{16,119,51}

\def\taskFullName{{Transformation Tracking and Understanding}\xspace}
\def\taskShortName{\textsc{TTU}\xspace}
\def\methodFullName{{Tracking Instances via Proposals and Overlaps}\xspace}
\def\methodName{\textsc{TIPO}\xspace}
\def\datasetName{\textsc{VISTA}\xspace}
\def\datasetFullName{Video Instances, States, and Transformation Annotations\xspace}

\definecolor{cvprblue}{rgb}{0.21,0.49,0.74}
\usepackage[pagebackref,breaklinks,colorlinks,allcolors=cvprblue]{hyperref}

\edef\arxivrestore{%
  \textfloatsep=\the\textfloatsep\relax \dbltextfloatsep=\the\dbltextfloatsep\relax}
\def\confName{CVPR}
\def\confYear{2027}
\renewcommand{\dblfloatpagefraction}{0.6}

\title{Taking Apart and Putting Together:\\Tracking Objects Through Physical Transformations}

\author{
Davide Gardenal$^{1}$ \quad
Mingxuan Liu$^{1}$ \quad
Flavio Vella$^{1}$ \quad
Quentin Guimard$^{2}$ \quad
Elisa Ricci$^{1,3}$
\\[0.5em]
$^{1}$University of Trento\quad
$^{2}$AMIAD, Pôle Recherche, Palaiseau\quad
$^{3}$Fondazione Bruno Kessler
}

\begin{document}
\twocolumn[{%
\renewcommand\twocolumn[1][]{#1}%
\maketitle
\input{figures/teaser}\vspace{1em}%
}]

\begingroup\setlength{\parskip}{0pt}%
\begin{abstract}
We introduce \textit{\taskFullName} (\textit{\taskShortName}), extending video object tracking to environments where objects undergo physical assemblies and disassemblies.
\taskShortName requires continuously tracking objects through transformations, identifying new entities, maintaining a state graph of their relationships, and assigning semantic labels to transformations.
While foundation models (\eg, SAM~2/3) maintain tracklets through appearance changes, they cannot update semantic identities as objects physically evolve.
Prior work addresses only disassemblies, relying on expensive searches triggered only when tracking fails.
{We propose \textit{\methodFullName} (\textit{\methodName}), a training-free framework that continuously infers transformations via a \emph{generate--evaluate--judge} pipeline. \textit{\methodName} generates split and merge proposals from disjoint mask components, scores them against mask-decoder priors using inner and external overlap, and queries an MLLM on the top proposal for visual verification and semantic labeling.}
{To benchmark \taskShortName, we curate \textit{\datasetFullName} (\textit{\datasetName}), a dataset including state graphs with 124 assembly and 132 disassembly events across 50 sequences and over 26,000 frames.}
\methodName achieves state-of-the-art tracking and state-graph accuracy {on \datasetName{,} while operating $6\times$ to $15\times$} faster than\linebreak{} prior methods.\\
Project page: \href{https://davidegardenal.github.io/tipo-tracking}{davidegardenal.github.io/tipo-tracking}
\end{abstract}\par\endgroup

\vspace{-1em}

\input{sections/introduction}
\input{sections/related_work}
\input{sections/method_v3}
\input{sections/experiments}

\input{sections/conclusions}

\section*{Acknowledgments}
We acknowledge the CINECA award under the ISCRA initiative for the availability of high-performance computing resources and support. We also thank the OpenAI Researcher Access Program for providing API credits.

\section*{LLM Usage Statement}
We disclose the use of LLMs to assist the authors in proofreading the manuscript, drafting minor parts of the text, refining visualizations and drafting parts of the experiment code. LLMs were not used for research ideation.
The authors reviewed, tested, and verified all LLM-assisted content and take full responsibility for the contents of this paper.
The MLLMs used for the Judge and Labeler in \cref{sec:methodology} are part of the proposed method.

{
    \small
    \bibliographystyle{ieeenat_fullname}
    \bibliography{ref}
}

\clearpage\arxivrestore
\input{supp}

\end{document}

%% file: preamble.tex
\usepackage{cuted} %

\usepackage{currfile} %

\usepackage{caption} %

\usepackage{pgfplots}
\usepgfplotslibrary{groupplots}
\pgfplotsset{compat=1.18}
\definecolor{tipoDis}{HTML}{EB6834}
\definecolor{tipoAsm}{HTML}{1BAF7A}

%% file: prompt_box.tex
\usepackage{xcolor}
\usepackage{mdframed}
\usepackage{microtype}
\usepackage{inconsolata}

\definecolor{systembg}{RGB}{235, 240, 250}
\definecolor{systemborder}{RGB}{140, 160, 200}
\definecolor{userbg}{RGB}{240, 248, 240}
\definecolor{userborder}{RGB}{140, 190, 150}
\definecolor{assistantbg}{RGB}{255, 252, 240}
\definecolor{assistantborder}{RGB}{210, 180, 100}
\definecolor{rolelabel}{RGB}{80, 80, 90}

\newmdenv[
  skipabove=6pt, skipbelow=6pt,
  innerleftmargin=10pt, innerrightmargin=10pt,
  innertopmargin=8pt, innerbottommargin=8pt,
  linewidth=1pt,
  roundcorner=4pt,
]{promptbox}

\newenvironment{systemprompt}{%
  \begin{promptbox}[linecolor=systemborder, backgroundcolor=systembg]%
  \noindent{\footnotesize\sffamily\bfseries\color{rolelabel}SYSTEM}\par\smallskip
  \ttfamily\small
}{%
  \end{promptbox}%
}

\newenvironment{userprompt}{%
  \begin{promptbox}[linecolor=userborder, backgroundcolor=userbg]%
  \noindent{\footnotesize\sffamily\bfseries\color{rolelabel}USER}\par\smallskip
  \ttfamily\small
}{%
  \end{promptbox}%
}

%% file: figures/teaser.tex
\begin{center}
    \captionsetup{type=figure}
    \includegraphics[width=\textwidth]{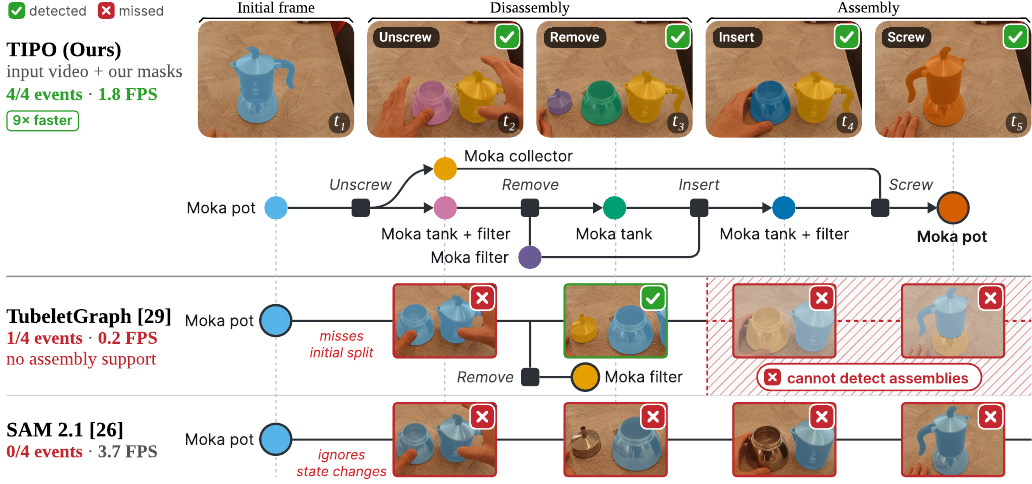}
    \caption{\textbf{\taskFullName.} Given a video and initial object prompts, \taskShortName requires tracking objects through assemblies and disassemblies while maintaining a part--whole state graph.
    Each row shows the masks and state graph across keyframes ($t_1$--$t_5$) for each method using the same SAM~2.1 backbone~\cite{ravi2025sam}.
    Foundation trackers (\eg, SAM~2/3~\cite{carion2026sam,ravi2025sam}) maintain pixel-level tracklets but ignore state changes.
    TubeletGraph~\cite{sun2026tracking} relies on reactive search after track loss, missing early splits ($t_2$) and failing on assemblies ($t_4, t_5$) at 0.2~FPS.
    \methodName proactively detects both events ($t_2$--$t_5$), tracking all parts and generating complete state graphs up to $9\times$ faster.}
    \label{fig:teaser}
\end{center}

%% file: sections/introduction.tex
\section{Introduction}

Imagine someone preparing a Moka pot (\cref{fig:teaser}, $t_1$): they unscrew the collector ($t_2$), pull out the funnel filter ($t_3$), and set the components aside.
In a matter of seconds, a single object splits into three distinct parts.
Modern visual foundation trackers (\eg, SAM~2~\cite{ravi2025sam}, SAM~3~\cite{carion2026sam}) are built around entity persistence: they follow objects through complex motion, but have no mechanism to revise identities when objects physically divide or merge.
As shown in the last row of \cref{fig:teaser} (SAM~2.1), the tracker either stubbornly preserves a single ID across a split ($t_2$) or arbitrarily drops fragmented parts ($t_3, t_4$).
Reassembling the pot ($t_4, t_5$) reverses the ambiguity: should the tracker follow separate components or recognize a unified \emph{Moka pot} again?
Yet the physical world is inherently dynamic~\cite{tokmakov2023breaking,xue2024learning}, with objects routinely assembled, disassembled, sliced, and folded~\cite{grauman2022ego4d,damen2022rescaling}.
Maintaining pixel tracklets is only half the challenge; dynamic environments require a system that \emph{tracks} objects through transformations, \emph{describes} the physical process, and \emph{updates} identities and part--whole relationships over time~\cite{zitkovich2023rt,mandikal2026mash}.

To address this dynamic scenario, we formalize the task of \textit{\taskFullName} (\taskShortName).
Given a video and initial object prompts, a \taskShortName system must (i)~track objects through assemblies and disassemblies, (ii)~describe the physical transformations, (iii)~update tracklet IDs and semantic labels, and (iv)~keep track of how every object relates to its parts over time.
The output is a single, unified state graph paired with object masks (\cref{fig:teaser}, {\methodName row}).
This structured representation naturally enables several downstream applications: robots can verify step completion before planning actions~\citep{intelligence2026pi}, and AR assistants can monitor current assembly states to guide users~\citep{plizzari2024outlook}.

Prior work only partially addresses this task.
TubeletGraph~\cite{sun2026tracking} handles isolated split events but has three major limitations for \taskShortName (\cref{fig:teaser}, {second row}).
First, its search for parts triggers only after track loss, missing the initial split ($t_2$) and detecting only later fragmentation ($t_3$).
Second, it is restricted to disassemblies and cannot model assemblies, causing identity switches ($t_4$) and failing to resolve assemblies ($t_5$).
Lastly, tracking dense candidate regions across the whole scene incurs severe overhead (0.2~FPS).

We take a fundamentally different approach.
Our key observation is simple: split and merge events are observable in the current frame's geometry.
In \cref{fig:teaser}, the collector's separation is instantly evident at $t_2$ because the tracked mask fragments into disconnected components.
There is no need to scan through the video to recover lost parts; the visual evidence is already in the frame.
Building on this, we introduce \textit{\methodFullName} (\methodName), a training-free framework for \taskShortName that explicitly models both assemblies and disassemblies on top of foundation trackers.

\input{figures/method_overview}
At each frame, \methodName executes a \emph{generate--evaluate--judge} pipeline.
It first \emph{generates} split and merge state proposals by partitioning the disjoint components of the tracked masks.
Next, it \emph{evaluates} each state proposal against promptable mask-decoder priors using complementary inner and external overlap metrics, distinguishing genuine transformations from visual occlusions without expensive search.
Finally, a Multimodal Large Language Model (MLLM)~\cite{qwen352026, gemma42026} \emph{judges} whether the top-scoring proposal matches the visual evidence in the frame and, if so, assigns semantic labels to the action and the newly formed objects (\eg, ``Remove'' and ``Moka filter'').
By querying the MLLM only on structurally validated proposals, \methodName continuously updates the state graph without compromising tracking speed.

\input{tables/data_summary}
To evaluate \taskShortName, we introduce \textit{\datasetFullName} (\datasetName), a challenging benchmark of 50 multi-object videos (26K+ frames) featuring dense state-graph annotations across 256 distinct transformations.
Unlike VOST-TAS~\cite{sun2026tracking}, which covers only disassemblies, \datasetName includes 124 assemblies and more than doubles the annotated transformations (\cref{tab:data_sum_horiz}).
Crucially, the state graphs enable a rigorous structural evaluation of each transformation, whereas VOST-TAS annotates only resulting parts,{ignoring which object transformed and rewarding methods that over-predict transformations (\cref{tab:vost_tas_hack}).}

\noindent \textbf{Contributions.} Our core contributions are: 
{\textit{i)} We formalize the \taskShortName task, extending prior tracking paradigms~\cite{sun2026tracking} to model both object assemblies and disassemblies.}
\textit{ii)} We propose \methodName, a training-free framework that replaces dense spatiotemporal search~\cite{sun2026tracking} with per-frame geometric evaluation of state proposals using mask-decoder priors, backed by sparse MLLM validation and semantic labeling.
{\textit{iii)} We curate \datasetName, the first benchmark to provide comprehensive state-graph annotations for both assembly and disassembly.}
\textit{iv)} We show through extensive experiments that \methodName establishes a new state-of-the-art for tracking and state-graph accuracy, while running {$6\times$ to $15\times$} faster than competitors.

%% file: figures/method_overview.tex
\begin{figure*}[t]
    \centering
    \includegraphics[page=1, width=\textwidth]{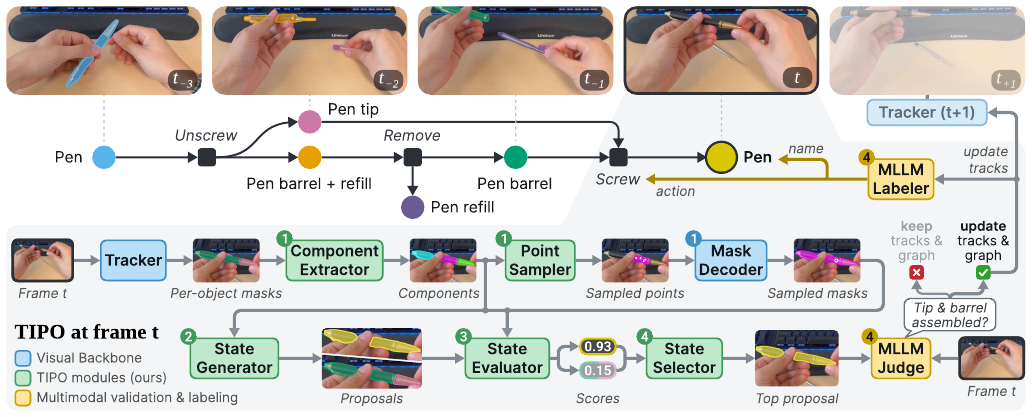}
    \caption{\textbf{Overview of \methodName.} At each frame $I_t$, the tracker provides the set of per-object masks. The pipeline operates in four stages \textbf{(1--4)}: \textbf{(1)} extraction of the disjoint components and mask sampling, \textbf{(2)} generation of all possible assembly and disassembly states, \textbf{(3)} evaluation of the state proposals based on mask overlap metrics, \textbf{(4)} validation of the highest-scoring proposal by the MLLM Judge and semantic labeling. Lastly, the state graph and the tracker are updated with the validated state.}

\label{fig:method_overview}
\end{figure*}

%% file: tables/data_summary.tex
\begin{table}[t]
\centering
\caption{\textbf{Comparison between \datasetName and VOST-TAS datasets}. \datasetName adds complex assembly events, doubles the number of transformations, and includes longer video sequences.}
\label{tab:data_sum_horiz}

\footnotesize
\setlength{\tabcolsep}{3pt}

\begin{tabular*}{\linewidth}{@{\extracolsep{\fill}}lccccccc@{}}
\toprule
\multirow{2}{*}{Dataset} & \multirow{2}{*}{Videos} & \multirow{2}{*}{Frames} & \multirow{2}{*}{Len. (s)} & \multirow{2}{*}{FPS} & \multicolumn{3}{c}{Transformations} \\
\cmidrule(l){6-8}
 & & & & & Asm. & Dis. & Total \\
\midrule
VOST-TAS \cite{sun2026tracking} & \textbf{57} & 7,820 & 22.3 & 5 & 0 & 108 & 108 \\
\datasetName (Ours) & 50 & \textbf{26,470} & \textbf{45.0} & \textbf{10--20} & \textbf{124} & \textbf{132} & \textbf{256} \\
\bottomrule
\end{tabular*}
\end{table}

%% file: sections/related_work.tex
\section{Related Work}
\noindent \textbf{Object Tracking and Foundation Models.}
Traditional multi-object tracking and video object segmentation primarily rely on tracking-by-detection \cite{bewley2016simple, zhang2022bytetrack, cao2023observation} and tracking-by-query \cite{sun2020transtrack, zeng2022motr, meinhardt2022trackformer} paradigms.
While accurate, these methods often struggle with target deformation and scale variation, and require extensive annotated training data \cite{yang2023track, cheng2023segment}.
To address these issues, recent work leverages foundation models such as SAM~\cite{kirillov2023segment} for zero-shot instance association~\cite{li2024matching}, interactive tracking~\cite{yang2023track, cheng2023segment}, and unified real-time tracking with streaming memory in SAM 2~\cite{ravi2025sam}. SAM 3 further extends this line to Promptable Concept Segmentation by decoupling recognition and localization~\cite{carion2026sam}.
However, these models still assume structural integrity: identity tracking breaks down when objects fragment or merge. While recent human-in-the-loop systems mitigate this via online correction~\cite{yang2026live}, they remain annotation tools rather than autonomous models. In contrast, we explicitly model structural transformations.
By repurposing SAM's zero-shot segmentation priors, \ie sampling internal mask representations, we isolate the structural cues needed to autonomously track split and merge events, eliminating the need for human intervention or domain-specific fine-tuning.

\noindent \textbf{Understanding Object Transformations.}
Recent literature explores physical state changes in videos.
Early work targeted human-driven events in egocentric datasets \cite{grauman2022ego4d, damen2022rescaling}, introducing benchmarks like VOST~\cite{tokmakov2023breaking} and VSCOS~\cite{yu2023video}.
Subsequent work pursued fine-grained understanding: VidOSC temporally localizes open-world transformations \cite{xue2024learning}, while SPOC spatially segments progressive changes \cite{mandikal2026spoc} for robotic manipulation \cite{mandikal2026mash}.
Most recently, Sun \etal \cite{sun2026tracking} bridged state recognition and tracking via the \emph{Track Any State} task and VOST-TAS benchmark. Their TubeletGraph framework {tracks all candidate regions of the scene as tubelets and, after a transformation, filters them with semantic and spatial priors to decide which belong to the target.} However, {tracking the whole scene} is computationally heavy, and {the framework is} strictly limited to disassembly events.
In contrast, \methodName natively models both assemblies and disassemblies. By evaluating state proposals via per-frame spatial overlap, we achieve accurate transformation tracking at a fraction of the computational cost.

\noindent \textbf{MLLMs for Understanding Object Transformations.}
Recent MLLMs, including both closed-source models \cite{gpt542026, claude47opus2026, gemini31pro2026} and open-weight architectures \cite{qwen352026, gemma42026}, have shown strong capabilities in parsing long-context video, enabling the grounding of dynamic physical properties, causality, and structural changes \cite{krojer2025a, zhou2026pai}.
For tracking transformations, prior work primarily used MLLMs as offline pseudo-label generators for training \cite{xue2024learning} or as post-hoc captioners to extract textual descriptions from mask-contoured keyframes \cite{sun2026tracking}.
In contrast, our pipeline integrates an MLLM to validate state changes directly within the tracking loop, effectively filtering out implausible false positives.

%% file: sections/method_v3.tex
\section{Methodology}
\label{sec:methodology}

\subsection{Problem Formulation}
\label{sec:problem_formulation}
We formalize the task of \textit{\taskFullName}.
Given a video $\mathcal{V} = \{I_t\}_{t=0}^{T}$ and a set of initial objects of interest $\mathcal{O}_0$ defined by their initial masks provided by the user $\mathcal{M}_0=\{m_0^i\}_{i\in\mathcal{O}_0}$, the goal is to continuously track and segment the objects, detect physical transformations, and output a directed acyclic \textit{state graph} $\mathcal{G} = (V, E)$ modeling their spatiotemporal and semantic evolution.
Unlike single-object tracking formulations that assume all resulting entities derive from a single progenitor, \taskShortName naturally models arbitrary multi-object interactions, where distinct entities can independently persist, split into parts, or merge to form new objects.

A node $v^i_t = (t, i, m^i_t, n) \in V$ represents the complete state of object $i$ at frame $t$, comprising its binary mask $m^i_t$ and a free-form text name $n$ (\eg, ``Pen''). An edge $e = (v^j_{t-1}, v^i_{t}, d) \in E$ captures the temporal transition from frame $t-1$ to $t$. Each edge is annotated with a textual description $d$ of the transformation (\eg, ``Unscrew''), or defaults to \textsc{none} when the object undergoes purely spatial motion without a state change. Edges strictly connect temporally adjacent frames. Detailed structural constraints governing the state graph are provided in Appendix~\ref{apdx:formulation}.

We characterize the two primary types of transformations by the connectivity of $\mathcal{G}$. An \textbf{\textit{assembly}} occurs when multiple components merge to form a new unified entity. Formally, a node $v^a_t \in V$ represents a newly assembled object if its in-degree from the previous frame is at least two: 
\begin{equation}
    |\{e = (v^i_{t-1}, v^a_t, d) \in E \mid v^i_{t-1} \in V\}| \ge 2
\end{equation}
Conversely, a \textbf{\textit{disassembly}} occurs when a single entity splits into constituent parts. A node $v^b_{t-1} \in V$ undergoes disassembly if its out-degree to the next frame is at least two:
\begin{equation}
    |\{e = (v^b_{t-1}, v^i_t, d) \in E \mid v^i_t \in V\}| \ge 2
\end{equation}

\subsection{Our Approach: \methodFullName}
\label{sec:methodology_framework}

\paragraph{Overview.} 
We introduce \methodFullName. As shown in \cref{fig:method_overview}, \methodName is a modular framework built on top of pre-trained segmentation trackers and MLLMs, requiring no additional training.

We initialize the tracker and state graph $\mathcal{G}$ using the initial objects $\mathcal{O}_0$. At each frame $I_t$, the tracker extracts the \textit{per-object masks} $\mathcal{M}_t=\{m^i_t\}_{i\in\mathcal{O}_t}$.
\methodName processes these masks through four interconnected stages:
\textbf{(1) Component Extraction \& Object Mask Sampling} decomposes the \textit{per-object masks} into \textit{spatially disjoint components} and prompts a mask decoder to extract per-component \textit{sampled masks}.
\textbf{(2) State Generation} enumerates plausible combinations of the \textit{spatially disjoint components} (for disassemblies) and \textit{sampled masks} (for assemblies) to output a set of \textit{state proposals}.
\textbf{(3) State Evaluation} scores the \textit{state proposals} using spatial overlap metrics applied to the \textit{sampled masks}, yielding \textit{evaluated state scores}.
Finally, \textbf{(4) State Selection, MLLM Validation \& Graph Update} filters the \textit{evaluated state scores} to query an MLLM, visually verifying the top-scoring state proposal. If validated, the framework assigns semantic labels and outputs the \textit{updated state graph} $\mathcal{G}$.

\paragraph{(1) Component Extraction \& Object Mask Sampling.}
\label{sec:comp_extr_samp}
At frame $I_t$, the raw masks $\mathcal{M}_t$ may be structurally ambiguous: a mask might fragment due to true disassembly or temporary occlusion, while multiple masks might fail to encompass a new assembly.
To resolve this, the Component Extractor first decomposes each tracked mask $m^i_t$ into spatially disjoint components $\mathcal{C}^i = \{c^i_0, \dots, c^i_j\}$ via Parallel Union-Find~\cite{bolelli2024state}. Hereafter, we omit the temporal index $t$.

Next, we leverage promptable mask decoders (\eg, SAM~\cite{kirillov2023segment}). For each component $c^i_j \in \mathcal{C}^i$, the Point Sampler outputs random points to prompt the Mask Decoder. To mitigate sampling variance, we ensemble multiple queries to extract a robust, zero-shot sampled mask $s_j^i$ (see Appendix~\ref{apdx:impl_details}). This distinguishes genuine transformations from visual artifacts: for disassemblies, the decoded mask of a merely occluded part still reconstructs the entire object; for assemblies, merged objects yield heavily overlapping sampled masks. We define the set of sampled masks for an arbitrary group of components $C$ as $S_C = \{s_j \mid c_j \in C\}$, the full sampled mask of object $i$ as $s^i = \bigcup_{s_j \in S_{\mathcal{C}^i}}s_j$, and the sampled masks for an object group $O$ as $S_O = \{s^i \mid i \in O\}$.

\paragraph{(2) State Generation.}
Using the spatially disjoint components $\mathcal{C}^i$ and the sampled masks extracted in the previous stage, we enumerate all plausible physical transformations. We define the set of all possible disassembly state proposals for object $i$ as the partition set of its components $\mathcal{C}^i$:
\begin{equation}
    \mathcal{D}^i = \Pi(\mathcal{C}^i) = \{h^i_0, \dots, h^i_p\}, \quad \text{with } h^i_j = \{C_0, \dots, C_k\}
\end{equation}
where each group $C \subseteq \mathcal{C}^i$ contains components proposed to form a single part. Similarly, the set of assembly state proposals is the partition set of tracked objects $\mathcal{O}_t$:
\begin{equation}
    \mathcal{A} = \Pi(\mathcal{O}_t) = \{ g_0, \dots, g_q\}, \quad \text{with } g_j = \{O_0, \dots, O_m\}
\end{equation}
These comprehensive sets of state proposals ($\mathcal{D}^i$ and $\mathcal{A}$) are outputted to the evaluator. In Appendix~\ref{apdx:impl_details} we formulate computational bounds that limit the cardinality of these sets.

\paragraph{(3) State Evaluation.}
\label{sec:state_eval}
To rank the generated disassembly ($h \in \mathcal{D}^i$) and assembly ($g \in \mathcal{A}$) state proposals, we define the spatial overlap scores
\begin{equation}
\begin{aligned}
    D(h) &= \prod_{C_n \in h} \Phi_{\text{in}}(C_n) \cdot (1 - \Phi_{\text{ex}}(C_n)) \\
    A(g) &= \prod_{O_n \in g} \Psi_{\text{in}}(O_n) \cdot (1 - \Psi_{\text{ex}}(O_n))
\end{aligned}
\end{equation}
where $\Phi$ and $\Psi$ denote the recall-based scoring functions for disassemblies and assemblies, respectively (formally defined below).
Intuitively, in the disassembly case, a high \textit{inner recall} $\Phi_{\text{in}}$ indicates that all components in $C_n$ belong to a single physical object and are thus correctly grouped, so $D(h)$ rewards it. Conversely, a high \textit{external recall} $\Phi_{\text{ex}}$ indicates that some components in $C_n$ overlap with those of other groups, signaling an incorrect grouping, so $D(h)$ penalizes it. The same reasoning applies to the assembly case, with $\Psi_{\text{in}}$ and $\Psi_{\text{ex}}$ evaluated on object groups $O_n$ instead of component groups $C_n$.

We define inner and external recall in terms of the mask recall ${R(a, b) = \lvert a \cap b \rvert / \lvert a \rvert}$, measuring the proportion of mask $a$'s area overlapped by mask $b$.

\textbf{Inner Recall} measures whether the sampled mask of each element in a group ($S_{C_n}, S_{O_n}$) overlaps all other elements of the same group:
\begin{equation}
\label{eq:inner}
\begin{aligned}
    \Phi_{\text{in}}(C_n) &= \frac{1}{|C_n|}\sum_{s_i \in S_{C_n}} R(\cup_{c_j \in C_n} c_j, s_i) \\
    \Psi_{\text{in}}(O_n) &= \frac{1}{|O_n|}\sum_{s_i \in S_{O_n}} \prod_{s_j \in S_{O_n}} R(s_i, s_j)
\end{aligned}
\end{equation}
For disassembly, $\Phi_{\text{in}}$ computes, for each sampled mask $s_i \in S_{C_n}$, the proportion of the group mask ($\cup_{c_j \in C_n} c_j$) covered by $s_i$. Comparing against the union, rather than against each component individually, prevents small outlier components from biasing the score.
For assembly, $\Psi_{\text{in}}$ instead computes, for each sampled mask $s_i \in S_{O_n}$, the product of its recalls with every sampled mask $s_j$ in the group. This requires the mask decoder to agree on the physical object regardless of which object is prompted, since the contribution of $s_i$ drops to zero as soon as it fails to overlap even a single $s_j$.
In both cases, the per-mask values are averaged over the group, so a high inner recall indicates that each sampled mask reconstructs the entire group, suggesting that all grouped elements belong to the same physical object.

\textbf{External Recall} measures how much the sampled masks of a group ($S_{C_n}, S_{O_n}$) leak into elements outside the group:
\begin{equation}
\label{eq:extern}
\begin{aligned}
    \Phi_{\text{ex}}(C_n) &= R(\cup_{c_k \in \mathcal{C}^i \setminus C_n} c_k, \cup_{s_i \in S_{C_n}} s_i) \\
    \Psi_{\text{ex}}(O_n) &= R(\cup_{s_e \in S_{\bar{O}_n}} s_e, \cup_{s_i \in S_{O_n}} s_i)
\end{aligned}
\end{equation}
where $\mathcal{C}^i \setminus C_n$ is the set of components of object $i$ excluding $C_n$, and $\bar{O}_n = \mathcal{O}_t \setminus O_n$ denotes the set of all tracked objects at frame $t$ not included in $O_n$. 
For disassembly, $\Phi_{\text{ex}}$ computes the proportion of the area of all external components ($\cup_{c_k \in \mathcal{C}^i \setminus C_n} c_k$) overlapped by all the sampled masks of the group ($\cup_{s_i \in S_{C_n}} s_i$).
For assembly, $\Psi_{\text{ex}}$ analogously computes the proportion of the area of all external sampled masks ($\cup_{s_e \in S_{\bar{O}_n}} s_e$) covered by all the sampled masks of the group ($\cup_{s_i \in S_{O_n}} s_i$).
In both cases, a high external recall indicates that the sampled masks of the group extend over elements assigned to other groups, signaling a leakage and thus an incorrect grouping.

Notably, in both inner and external recall, disassembly compares against the \textit{actual tracks}, whereas assembly compares against the \textit{sampled masks}. When objects assemble, the tracker may propagate each track to only a portion of the newly merged object, and comparing against such partial tracks would make both metrics unreliable. The sampled masks, obtained by prompting the mask decoder, instead recover the full extent of each object.
Finally, the proposals and their scores $D(h)$ and $A(g)$ are passed to the selection and validation stage.

\paragraph{(4) State Selection, MLLM Validation \& Graph Update.}
\label{sec:state_selec}

We filter evaluated state proposals by a threshold $\tau$, extracting the top-scoring disassembly state $\hat{h}^i = \arg\max_{h \in \mathcal{D}^i} D(h)$ and assembly state $\hat{g} = \arg\max_{g \in \mathcal{A}} A(g)$. If no proposal exceeds $\tau$, a default no-assembly and no-disassembly state is selected. 

Because random mask sampling can yield high-scoring false positives, an MLLM Judge independently validates the selected proposals. The MLLM receives sparsely sampled historical frames (\eg, 5s, 3s, and 1s prior) alongside the current frame $I_t$ annotated with the proposed regions to visually verify if the proposed transformation actually occurred. Positioning the MLLM downstream of our overlap metrics limits the number of expensive queries.

Upon validation, we update the tracker and the state graph $\mathcal{G}$, instantiating new nodes and edges with unique tracking IDs. To assign semantic labels, an MLLM Labeler processes a historical frame, the annotated frame $I_t$, and a list of involved components. We parse its structured JSON response to populate node names $n$ (\eg, ``Pen'') and edge descriptions $d$ (\eg, ``Screw''). These masks then condition the tracker for the next frame. Appendix~\ref{apdx:impl_details} details all MLLM prompts.

\subsection{The \datasetName Benchmark}
\label{sec:dataset}
\input{tables/tuvos_summary}
\paragraph{Overview.} 
While prior benchmarks like VOST-TAS~\cite{sun2026tracking} consider state changes, they annotate only the resulting parts rather than modeling the transformation process itself. To address this gap and rigorously evaluate \taskShortName, we introduce \textit{\datasetFullName} (\datasetName). As summarized in \cref{tab:data_sum_horiz}, \datasetName comprises 50 diverse sequences sourced from Assembly101 \cite{sener2022assembly101}, IKEA ASM \cite{ben2021ikea}, and custom recordings. It provides continuous, dense segmentation masks paired with comprehensive state graphs that explicitly link parts as they separate and merge over time. By modeling multi-stage assemblies and disassemblies and more than doubling the annotated transformations of prior work~\cite{sun2026tracking}, \datasetName provides a highly realistic testbed for dynamic environments (see {Appendix~\ref{apdx:composition} for the detailed dataset composition and }Appendix~\ref{apdx:annot} for the complete annotation protocol).

\noindent \textbf{Evaluation Metrics.} 
We perform evaluations across two primary dimensions: mask tracking and state-graph generation. We describe the metrics below (details and formal definitions in Appendix~\ref{apdx:eval_metrics}).

\textsc{Tracking Quality}. Following established standards~\cite{tokmakov2023breaking}, we report the average mask Intersection-over-Union (IoU) $\mathcal{J}$ across the full sequence. To measure track survival after transformations, we report $\mathcal{J}_{tr}$, computed exclusively on the last 25\% of frames.

\textsc{Transformation Tracking}. 
 Leveraging our dense annotations, we propose $\mathcal{T}_{F_1}$ to comprehensively assess the structural state graph via bipartite matching. A match is considered valid only if it achieves: (1) identical input and output object counts, (2) correct temporal boundaries, and (3) object-level IoU $ > 0.5$ across pre- and post-transformation masks. We report the macro-averaged $F_1$ score, whose precision $\mathcal{T}_p$ and recall $\mathcal{T}_r$ count only the matches satisfying all three conditions, unlike prior metrics (see discussion below).

\textsc{Transformation Semantics}. Following prior work~\cite{sun2026tracking}, we use an LLM-based evaluator to measure the open-vocabulary semantic similarity between predicted and ground-truth object names ($\mathcal{S}_O$) and action descriptions ($\mathcal{S}_A$). The overall semantic score $\mathcal{S}$ is their per-sequence harmonic mean, macro-averaged across all sequences.

\textsc{Overall Score}. As our primary metric for overall pipeline performance, $\mathcal{H}_{F_1}$ extends the structural bipartite matching of $\mathcal{T}_{F_1}$ to enforce semantic correctness, yielding a holistic $F_1$ score for the fully labeled state graph.

\input{tables/vost_tas_hack}
\input{tables/disassembly_only}
\noindent \textbf{Discussion.} 
Prior metrics~\cite{sun2026tracking} evaluate transformations through their temporal boundaries, using precision ($\mathcal{T}^{\text{TAS}}_P$) and recall ($\mathcal{T}^{\text{TAS}}_R$), and through the spatial overlap of the resulting objects, using $\mathcal{H}_{ST}$. These measures capture whether a transformation is detected at the correct time and whether its outputs are accurately localized. However, they do not directly assess whether those outputs are correctly associated with the input objects from which they originate. A prediction may therefore achieve accurate temporal boundaries and high output overlap while still linking the resulting objects to the wrong inputs, leaving the input-to-output transition itself unevaluated (see Appendix~\ref{apdx:tas_metrics}).
This is a serious weakness of previous evaluation frameworks, as even a simple baseline (AlwaysPredict) that predicts a disassembly at every mask split and an assembly at every mask overlap (with IoU $ > 0.5$) outperforms state-of-the-art methods such as TubeletGraph in $\mathcal{H}_{ST}$ on VOST-TAS by predicting $14\times$ to $15\times$ more transformations (\cref{tab:vost_tas_hack}).
{In contrast, our metrics leverage the dense annotations of \datasetName to verify the full predicted state graph, including the link between each transformation's input and output objects.}

%% file: tables/tuvos_summary.tex
\begin{table*}[t]
\centering
\footnotesize

\caption{\textbf{State-of-the-art comparison on \datasetName.} We evaluate pure tracking and state-graph generation using two different foundational trackers. Best results per tracker are in \textbf{bold}, and second-best are \underline{underlined}.}
\label{tab:tuvos_main}

\setlength{\tabcolsep}{4pt}
\renewcommand{\arraystretch}{1.05}

\begin{tabularx}{\linewidth}{
l
l
>{\centering\arraybackslash}X
>{\centering\arraybackslash}X
>{\centering\arraybackslash}X
>{\centering\arraybackslash}X
>{\centering\arraybackslash}X
>{\centering\arraybackslash}X
>{\centering\arraybackslash}X
>{\centering\arraybackslash}X
}
\toprule

\multirow{2}{*}{Tracker} &
\multirow{2}{*}{Method} &
\multicolumn{2}{c}{\textsc{Tracking}} &
\multicolumn{5}{c}{\textsc{State Graph}} &
\multirow{2}{*}{FPS $\uparrow$} \\

\cmidrule(lr){3-4}
\cmidrule(lr){5-9}

&
& $\mathcal{J}\uparrow$
& $\mathcal{J}_{tr}\uparrow$
& $\mathcal{S}\uparrow$
& {$\mathcal{T}_{r}\uparrow$}
& {$\mathcal{T}_{p}\uparrow$}
& $\mathcal{T}_{F_1}\uparrow$
& $\mathcal{H}_{F_1}\uparrow$
& \\

\midrule

\multirow{6}{*}{SAM 2.1}
& \blname{Vanilla SAM \cite{ravi2025sam}}
& \blname{83.1} & \blname{78.5}
& \blname{--} & {\blname{--}} & {\blname{--}} & \blname{--} & \blname{--}
& \blname{\textbf{3.7}} \\

& {\blname{AlwaysPredict}}
& {\blname{90.3}} & {\blname{\underline{87.2}}}
& {\blname{--}} & {\blname{25.3}} & {\blname{6.6}} & {\blname{9.1}} & {\blname{--}}
& {\blname{2.5}} \\

& {\blname{Agentic}}
& {\blname{77.5}} & {\blname{68.7}}
& {\blname{76.7}} & {\blname{7.6}} & {\blname{15.8}} & {\blname{9.9}} & {\blname{5.7}}
& {\blname{0.9}} \\

& TubeletGraph \cite{sun2026tracking}
& 88.1 & 83.4
& {36.3} & {9.1} & {23.2} & 12.5 & {2.6}
& 0.2 \\

&
\cellcolor{gray!10}TIPO \textit{w/o Judge}
& \cellcolor{gray!10}\underline{90.4}
& \cellcolor{gray!10}\underline{87.2}
& \cellcolor{gray!10}{76.8}
& \cellcolor{gray!10}\underline{{43.8}}
& \cellcolor{gray!10}\underline{{35.6}}
& \cellcolor{gray!10}\underline{36.7}
& \cellcolor{gray!10}\underline{{19.7}}
& \cellcolor{gray!10}\underline{3.0} \\

&
\cellcolor{gray!10}TIPO (Ours)
& \cellcolor{gray!10}\textbf{90.7}
& \cellcolor{gray!10}\textbf{88.6}
& \cellcolor{gray!10}\textbf{{78.9}}
& \cellcolor{gray!10}\textbf{{48.4}}
& \cellcolor{gray!10}\textbf{{46.7}}
& \cellcolor{gray!10}\textbf{44.7}
& \cellcolor{gray!10}\textbf{{27.1}}
& \cellcolor{gray!10}1.8 \\

\midrule

\multirow{6}{*}{SAM 3}
& \blname{Vanilla SAM \cite{carion2026sam}}
& \blname{87.9} & \blname{83.0}
& \blname{--} & {\blname{--}} & {\blname{--}} & \blname{--} & \blname{--}
& \blname{\textbf{6.1}} \\

& {\blname{AlwaysPredict}}
& {\blname{88.4}} & {\blname{83.4}}
& {\blname{--}} & {\blname{17.9}} & {\blname{8.2}} & {\blname{8.5}} & {\blname{--}}
& {\blname{3.8}} \\

& {\blname{Agentic}}
& {\blname{83.0}} & {\blname{75.5}}
& {\blname{67.5}} & {\blname{7.5}} & {\blname{19.7}} & {\blname{10.3}} & {\blname{4.9}}
& {\blname{1.0}} \\

& TubeletGraph \cite{sun2026tracking}
& 89.4 & 84.9
& {20.0} & {2.1} & {8.0} & 3.3 & {0.0}
& 0.4 \\

&
\cellcolor{gray!10}TIPO \textit{w/o Judge}
& \cellcolor{gray!10}\textbf{91.5}
& \cellcolor{gray!10}\textbf{88.0}
& \cellcolor{gray!10}{77.4}
& \cellcolor{gray!10}\underline{{37.3}}
& \cellcolor{gray!10}\underline{{37.5}}
& \cellcolor{gray!10}\underline{33.4}
& \cellcolor{gray!10}\underline{{16.8}}
& \cellcolor{gray!10}\underline{3.9} \\

&
\cellcolor{gray!10}TIPO (Ours)
& \cellcolor{gray!10}\underline{91.4}
& \cellcolor{gray!10}\underline{87.8}
& \cellcolor{gray!10}\textbf{{80.2}}
& \cellcolor{gray!10}\textbf{{44.7}}
& \cellcolor{gray!10}\textbf{{49.0}}
& \cellcolor{gray!10}\textbf{42.5}
& \cellcolor{gray!10}\textbf{{28.3}}
& \cellcolor{gray!10}2.4 \\

\bottomrule
\end{tabularx}
\end{table*}

%% file: tables/vost_tas_hack.tex
\begin{table}[t]
\centering
\footnotesize
\caption{\textbf{Over-predicting transformations on VOST-TAS.} The AlwaysPredict baseline predicts a disassembly at every mask split and an assembly at every mask overlap, resulting in $14\times$ to $15\times$ more transformations (\#Pred.) than TubeletGraph~\cite{sun2026tracking}, against 108 ground-truth transformations. The recall-based $\mathcal{T}^{\text{TAS}}_R$ and $\mathcal{H}_{ST}$ reward this behavior, while the temporal precision $\mathcal{T}^{\text{TAS}}_P$ verifies only the timing of each transformation, not the involved objects.}
\label{tab:vost_tas_hack}

\setlength{\tabcolsep}{3pt}
\begin{tabular*}{\linewidth}{@{\extracolsep{\fill}}llcccc@{}}
\toprule
Tracker & Method & \#Pred. & $\mathcal{T}^{\text{TAS}}_P\uparrow$ & $\mathcal{T}^{\text{TAS}}_R\uparrow$ & $\mathcal{H}_{ST}\uparrow$ \\
\midrule
\multirow{2}{*}{SAM 2.1}
& AlwaysPredict & 900 & 6.2 & 51.9 & 13.9 \\
& TubeletGraph & 65 & 38.4 & 23.1 & 13.0 \\
\midrule
\multirow{2}{*}{SAM 3}
& AlwaysPredict & 569 & 9.0 & 47.2 & 15.7 \\
& TubeletGraph & 38 & 29.0 & 10.2 & 6.5 \\
\bottomrule
\end{tabular*}
\end{table}

%% file: tables/disassembly_only.tex
\begin{table}[t]
\centering
\footnotesize
\caption{\textbf{Disassembly-only comparison on \datasetName.} We evaluate the methods on the disassembly-only split of VISTA to match the capabilities of TubeletGraph.
{We report results using two different foundational trackers. Best results per tracker are in \textbf{bold}, and second-best are \underline{underlined}.}}
\label{tab:disassembly_only}

\setlength{\tabcolsep}{4pt}
\renewcommand{\arraystretch}{1.05}

\begin{tabularx}{\linewidth}{
l
l
>{\centering\arraybackslash}X
>{\centering\arraybackslash}X
>{\centering\arraybackslash}X
}
\toprule

{Tracker} & Method & $\mathcal{T}_{r}\uparrow$ & $\mathcal{T}_{p}\uparrow$ & $\mathcal{T}_{F_1}\uparrow$ \\

\midrule

\multirow{3}{*}{{SAM 2.1}}
& {TubeletGraph \cite{sun2026tracking}}
& {22.7} & {33.7} & {25.6} \\

&
{\methodName \textit{w/o Judge}}
& {\textbf{42.4}}
& {\underline{36.7}}
& {\underline{37.5}} \\

&
{\methodName (Ours)}
& {\underline{42.1}}
& {\textbf{42.6}}
& {\textbf{39.8}} \\

\midrule

\multirow{3}{*}{{SAM 3}}
& TubeletGraph \cite{sun2026tracking}
& 3.5 & 9.5 & 4.9 \\

&
\methodName \textit{w/o Judge}
& \textbf{51.2}
& \underline{51.5}
& \underline{46.8} \\

&
\methodName (Ours)
& \underline{48.4}
& \textbf{60.4}
& \textbf{50.2} \\

\bottomrule
\end{tabularx}
\end{table}

%% file: sections/experiments.tex
\section{Experiments}

\input{figures/qualitatives}

{In this section, we compare \methodName with state-of-the-art methods on \datasetName, present qualitative results, and ablate its main components.}

\noindent \textbf{Experimental Setup.} 
To demonstrate that our framework is agnostic to the underlying foundation model, we evaluate \methodName using both SAM 2.1~\cite{ravi2025sam} and SAM 3~\cite{carion2026sam} as base trackers.
By default, we instantiate the MLLM Judge with Qwen 3.5 35B~\cite{qwen352026} and the MLLM Labeler with GPT-5.4~\cite{gpt542026}.
We evaluate two configurations: the full pipeline (\methodName), prioritizing accuracy, and a judge-free pipeline (\textit{\methodName w/o Judge}), prioritizing inference speed.
Appendix~\ref{apdx:impl_details} provides additional implementation details.

\subsection{Comparisons with State-of-the-art Methods}
\label{sec:main_results}
We benchmark \methodName against TubeletGraph~\cite{sun2026tracking} on our proposed \taskShortName task on \datasetName. {Since VOST-TAS annotates only the resulting parts of each transformation, $\mathcal{T}_{F_1}$ cannot be computed on it, and its TAS metrics reward over-predicting (\cref{tab:vost_tas_hack}). We therefore report the full VOST-TAS~\cite{sun2026tracking} comparison in Appendix~\ref{apdx:vost_tas_full}.} {We further compare against vanilla SAM and two baselines built on the same trackers. AlwaysPredict predicts transformations directly from tracker cues, without state generation and evaluation. The Agentic baseline gives the MLLM (Qwen 3.5 35B) direct control over the tracker through a tool-based harness with offline access to the full video (implem. details in Appendix~\ref{apdx:baselines}).}

\noindent\textbf{On \datasetName, \methodName achieves state-of-the-art transformation tracking while operating up to an order of magnitude faster.} 
As shown in \cref{tab:tuvos_main}, both \methodName configurations significantly outperform TubeletGraph across all metrics. By natively handling assemblies and avoiding heavy temporal search, \methodName improves tracking retention ($\mathcal{J}_{tr}$) by $+5.2$ points using SAM 2.1. This gap widens drastically in state-graph generation{, with} absolute {gains} of $+32.2$ $\mathcal{T}_{F_1}$ points with SAM 2.1 (44.7 vs 12.5) and $+39.2$ points with SAM 3 (42.5 vs 3.3). {\methodName also leads in both precision ($\mathcal{T}_p$) and recall ($\mathcal{T}_r$), showing that these gains do not come from trading one for the other.} {AlwaysPredict preserves strong tracking but reaches only 9.1 and 8.5 $\mathcal{T}_{F_1}$, $-27.6$ and $-24.9$ points below \methodName \textit{w/o Judge}, which likewise uses no MLLM to validate transformations. This demonstrates that tracker cues alone are insufficient to detect transformations. Notably, the Agentic baseline falls $-34.8$ and $-32.2$ $\mathcal{T}_{F_1}$ points behind \methodName with the same MLLM, degrades tracking below vanilla SAM, and runs $2.0\times$ and $2.4\times$ slower, despite its access to the full video.} Crucially, our framework delivers these state-of-the-art results while processing videos $6\times$ to $15\times$ faster than {TubeletGraph}, confirming that per-frame mask evaluation is vastly more efficient than dense tubelet search {(see Appendix~\ref{apdx:scalability} for a scalability analysis)}.

\noindent\textbf{{On disassemblies alone, \methodName still outperforms TubeletGraph.}}
{Since TubeletGraph cannot handle assemblies, we also evaluate on a split of \datasetName that removes them while keeping all 132 disassemblies (41 clips). As shown in \cref{tab:disassembly_only}, TubeletGraph doubles its $\mathcal{T}_{F_1}$ with SAM 2.1, from 12.5 on the full \datasetName to 25.6. Nevertheless, \methodName still outperforms it by $+14.2$ points, and by $+45.3$ points with SAM 3. Notably, even \methodName \textit{w/o Judge} retains gains of $+11.9$ and $+41.9$ points. This demonstrates that the advantage of \methodName stems from its per-frame proposal evaluation, and not only from its ability to handle assemblies.}

\noindent\textbf{{Continuous structural evaluation is robust to the choice of the foundation tracker.}}
{On \datasetName, upgrading TubeletGraph from SAM 2.1 to SAM 3 drops its $\mathcal{T}_{F_1}$ from 12.5 to 3.3, as its search is triggered only by track loss, which a stronger tracker makes rarer (Appendix~\ref{apdx:tg_sam3_failure}). In contrast, \methodName evaluates structural changes at every frame, rather than upon track loss, and achieves 44.7 and 42.5 $\mathcal{T}_{F_1}$ with SAM 2.1 and SAM 3, respectively.} {We further analyze the robustness of \methodName to tracker noise and its scaling with the tracker size in Appendix~\ref{apdx:robustness} and Appendix~\ref{apdx:tracker_size}, respectively.}

\noindent\textbf{\methodName unifies assemblies and disassemblies in complex, multi-step videos.}
{In} \cref{fig:quali_main}, the top row demonstrates a clean assembly sequence, while the bottom row showcases cascaded merges and splits natively unified into a consistent state graph. Additional qualitative results and limitations are discussed in Appendix~\ref{apdx:qualitatives}.

\input{tables/main_ablation}

\subsection{Ablation Study}
\label{sec:ablation_study}

{
\noindent\textbf{Single-point prompts let SAM resolve ambiguity through its own prior.} A single point is inherently ambiguous: it may belong to a sub-part, a part, or the whole object. SAM resolves this by selecting the most plausible mask, revealing which object it believes the point belongs to. Ensembling these single-point decodings probes SAM's zero-shot objectness prior without biasing the decoding. Adding points removes this ambiguity: the decoder must cover all prompts, and the mask collapses onto the sampled component. For an occluded object, the decoded mask no longer extends to other fragments, producing spurious disassemblies; for assemblies, merged parts no longer yield overlapping masks. As shown in \cref{tab:sampled_points_ablation}, sampling $K=4$ points drops $\mathcal{T}_{F_1}$ by 11.7 with the MLLM Judge and 11.3 without it.
}

\noindent\textbf{Both inner and external mask overlaps are strictly required to evaluate state proposals.} In \cref{eq:inner,eq:extern}, we formulated $\Phi_{\text{in}}, \Psi_\text{in}$ and $\Phi_{\text{ex}}, \Psi_\text{ex}$ as the foundational metrics for assessing assemblies and disassemblies. \cref{tab:formula_ablation} demonstrates the critical necessity of external recall ($\Phi_{\text{ex}}, \Psi_\text{ex}$). Excluding these terms mathematically blinds the evaluation to whether a sub-state incorrectly bleeds into external components, causing performance to collapse by $-17.7$ $\mathcal{T}_{F_1}$ points for assemblies and $-15.3$ $\mathcal{T}_{F_1}$ points for disassemblies. {For instance, a hand holding a pen splits its mask into two components $c_1, c_2$ whose sampled masks both cover the whole pen. Inner recall alone then scores the spurious split $\{\{c_1\},\{c_2\}\}$ as high as the intact pen $\{\{c_1,c_2\}\}$, whereas $\Phi_{\text{ex}}$ penalizes the leakage of each part into the other.}

\noindent\textbf{Advanced reasoning models provide precise transformation classification.} We evaluate the impact of the MLLM Labeler in populating the final state-graph labels. As shown in \cref{tab:vlm_sem_abl}, while object naming accuracy ($\mathcal{S}_O$) remains relatively stable across most models, action naming ($\mathcal{S}_A$) requires deeper temporal understanding. GPT-5.4 and GPT-5.4 mini outperform the best local model by at least $+9.8$ $\mathcal{S}_A$ points, while GPT-5.4 delivers the highest overall semantic accuracy $\mathcal{S}$ and is our default MLLM Labeler.

\noindent\textbf{The MLLM Judge substantially improves spatial reasoning, though the Judge-free pipeline remains highly competitive.} The MLLM Judge acts as a final filter to verify whether the highest-scoring state proposal actually occurred.
As shown in {\cref{tab:tuvos_main}}, integrating the judge yields a definitive improvement of $+8.0$ to $+9.1$ $\mathcal{T}_{F_1}$ points across trackers.
However, achieving over $33.0$ $\mathcal{T}_{F_1}$ relying solely on mask overlap scoring shows that the Judge-free configuration suits deployments with strict computational constraints.

%% file: figures/qualitatives.tex
\begin{figure*}[t]
    \centering
\includegraphics[width=\textwidth]{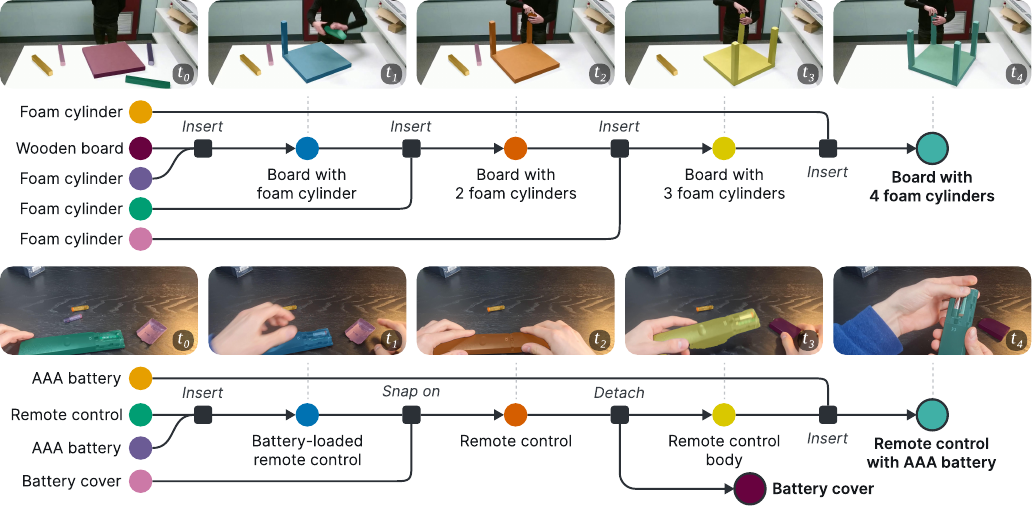}
    \caption{\textbf{Qualitative results on \datasetName}. Visualizing the \methodName pipeline on cascaded transformations. The state graphs map part-whole interactions and semantic actions, while the masks show robust spatial continuity across multi-step assembly and disassembly sequences.}
    \label{fig:quali_main}
\end{figure*}

%% file: tables/main_ablation.tex
\begin{table}[t]
    \centering
    \footnotesize

    \caption{\textbf{Validating the design of \methodName.} Experiments are conducted on \datasetName using SAM 3 as the tracker and Qwen 3.5 35B as the MLLM Judge. Minimal point sampling prevents spatial over-conditioning on local sub-parts, {and} external mask recall for both assembly ($\Psi$) and disassembly ($\Phi$) state evaluation is essential to penalize incorrect grouping.}
    \label{tab:main_vlm_ablations}
    \label{tab:sampled_points_ablation}
    \label{tab:formula_ablation}

    \begin{tabular*}{\linewidth}{@{\hspace{4pt}\extracolsep{\fill}}ccccccc@{\hspace{4pt}}}
        \toprule
        \multirow{2}{*}{\shortstack{MLLM\\Judge}} & \multirow{2}{*}{Pts} & \multicolumn{2}{c}{Ext. recall} & \multirow{2}{*}{$\mathcal{T}_{r}\uparrow$} & \multirow{2}{*}{$\mathcal{T}_{p}\uparrow$} & \multirow{2}{*}{$\mathcal{T}_{F_1}\uparrow$} \\
        \cmidrule(lr){3-4}
         & & $\Psi$ & $\Phi$ & & & \\
        \midrule
        \multirow{2}{*}{\cmark} & 1 & \cmark & \cmark & \textbf{44.7} & \textbf{49.0} & \textbf{42.5} \\
                                & 4 & \cmark & \cmark & 31.4 & 38.3 & 30.8 \\
        \midrule
        \multirow{4}{*}{\xmark} & 1 & \cmark & \cmark & \textbf{37.3} & \textbf{37.5} & \textbf{33.4} \\
                                & 4 & \cmark & \cmark & 23.1 & 26.8 & 22.1 \\
                                & 1 & \xmark & \cmark & 12.5 & 28.0 & 15.7 \\
                                & 1 & \cmark & \xmark & 15.7 & 23.4 & 18.1 \\
        \bottomrule
    \end{tabular*}
\end{table}

\begin{table}[t]
    \centering
    \footnotesize

    \caption{\textbf{Semantic graph labeling.} Advanced reasoning models (\eg, GPT-5.4) are more precise at action naming ($\mathcal{S}_A$), but not at object naming ($\mathcal{S}_O$). {Experiments are conducted on \datasetName using SAM 3 as the tracker.}}
    \label{tab:vlm_sem_abl}

    \begin{tabular*}{\linewidth}{@{\hspace{4pt}\extracolsep{\fill}}lcccc@{\hspace{4pt}}}
        \toprule
        MLLM Labeler & Local & $\mathcal{S}_A\uparrow$ & $\mathcal{S}_O\uparrow$ & $\mathcal{S}\uparrow$ \\
        \midrule
        Qwen 3.5 35B & \multirow{2}{*}{\cmark} & {80.2} & {80.0} & {73.4} \\
        Gemma 4 31B  &                         & {78.3} & {82.2} & {72.8} \\
        \midrule
        GPT-5.4      & \multirow{2}{*}{\xmark} & {90.0} & {82.0} & {80.2} \\
        GPT-5.4 mini &                         & {92.6} & {73.8} & {76.7} \\
        \bottomrule
    \end{tabular*}
\end{table}

%% file: sections/conclusions.tex
\section{Conclusion}

We introduced \taskFullName to generalize video tracking to environments where objects undergo structural changes, unifying assembly and disassembly into continuous state graphs. To benchmark this task, we curated \datasetName, featuring 256 densely annotated, multi-step transformations. We also proposed \methodName, a training-free framework that replaces reactive spatiotemporal search with continuous, per-frame structural evaluation and MLLM validation. By leveraging the zero-shot priors of modern foundation models, \methodName achieves state-of-the-art {tracking and state-graph accuracy}, operating up to an order of magnitude faster than prior work. Our results demonstrate the effectiveness of continuous structural modeling for tracking objects in complex, real-world environments.

%% file: supp.tex
\clearpage
\setcounter{page}{1}
\maketitlesupplementary
\appendix
\renewcommand{\dblfloatpagefraction}{0.5}
\input{appendix/appendix_overview}
\input{appendix/extended_formulation}
\input{appendix/extended_method_implem}
\input{appendix/dataset}
\input{appendix/additional_quantitative}
\input{appendix/extended_comparisons}
\input{appendix/additional_qualitative}

%% file: appendix/appendix_overview.tex
\section*{Appendix Overview}
\label{sec:appendix_overview}
This supplementary material provides technical details, extended evaluations, and qualitative results to support the main manuscript. The contents are organized as follows:

\vspace{1em}

\noindent\textbf{Appendix~\ref{apdx:formulation}: Extended Problem Formulation.} 
Details the formal graph constraints and formulations for the \taskShortName task.

\vspace{0.5em}
\noindent\textbf{Appendix~\ref{apdx:impl_details}: Extended Methodology and Implementation.} 
Provides the ensembled mask sampling protocol, algorithmic bounds for state generation, exact MLLM prompts, hyperparameters, and {software and} hardware specifications required to reproduce \methodName.

\vspace{0.5em}
\noindent\textbf{Appendix~\ref{apdx:dataset}: The \datasetName Benchmark.} 
Contains expanded details on {the dataset composition,} the annotation protocol, the custom annotation interface, the formal mathematical evaluation metrics{, and a discussion of identity switches}.

\vspace{0.5em}
\noindent\textbf{Appendix~\ref{apdx:extra_quant}: Additional Quantitative Analysis.} 
Provides expanded ablations on hyperparameter sensitivity, MLLM model selection, {tracker size}, a multi-seed variability analysis {on \datasetName}{, a runtime scalability analysis, a robustness analysis on tracker noise, and an error propagation analysis}.

{\vspace{0.5em}
\noindent\textbf{Appendix~\ref{apdx:extended}: Extended Comparisons.}
Provides {two additional baselines for \taskShortName, one using only the tracker and one with an MLLM agent controlling the tracker through tools,} results with a fully open-weight pipeline, {the full comparison on VOST-TAS,} and a tracking-only comparison with state-of-the-art VOS models on VOST.}

\vspace{0.5em}
\noindent\textbf{Appendix ~\ref{apdx:qualitatives}: Qualitative Results and Failure Analysis.} 
Provides {a non-rigid transformation example,} extensive visual sequences{ of} complex assemblies{ and failure cases}, and a transparent discussion of current limitations.

\vspace{1.5em}
\hrule
\vspace{1.5em}

%% file: appendix/extended_formulation.tex
\section{Extended Problem Formulation}
\label{apdx:formulation}

Expanding upon the task definition of \textit{\taskFullName} presented in \cref{sec:problem_formulation}, we formally detail the structural constraints that ensure the state graph $\mathcal{G} = (V, E)$ maintains structural and temporal consistency. 

\vspace{0.5em}
\noindent\textbf{Temporal Adjacency and Directionality.} 
To strictly enforce natural temporal flow, every edge must connect states across exactly one forward time step, rendering the graph a Directed Acyclic Graph (DAG). For any edge connecting two nodes, the temporal indices must be consecutive:
\begin{equation}
    \forall e = (v^j_{t_a}, v^i_{t_b}, d) \in E \implies t_b = t_a + 1
\end{equation}

\vspace{0.5em}
\noindent\textbf{Identity Preservation (Spatial Motion).} 
If an object undergoes pure spatial motion or occlusion without a physical state change, its transformation description defaults to \textsc{none}. In this scenario, the object's unique tracking identifier must be strictly preserved across the edge:
\begin{equation}
    \forall e = (v^j_{t}, v^i_{t+1}, \textsc{none}) \in E \implies j = i
\end{equation}

\vspace{0.5em}
\noindent\textbf{Identity Assignment after Transformations.} 
After any assembly or disassembly, the nature of the affected entities is fundamentally altered. Consequently, the resulting pre- and post-transformation nodes must possess entirely distinct identifiers. Let $d$ be any action description:
\begin{equation}
\begin{aligned}
    &\forall e = (v^j_{t}, v^i_{t+1}, d) \in E \quad \text{where}\quad d \neq \textsc{none} \\
    &\qquad \implies j \neq i
\end{aligned}
\end{equation}
Furthermore, to prevent duplicate identities coexisting in the physical space, no two distinct nodes within the same frame can share an object identifier:
\begin{equation}
    \forall v^j_t, v^i_t \in V \quad\text{where} \quad v^j_t \neq v^i_t \implies j \neq i
\end{equation}

\vspace{0.5em}
\noindent\textbf{Semantic Consistency.}
All the descriptions $d$ in the set of edges $E_T\subseteq E$ representing a transformation must be equal to preserve semantic consistency:
\begin{equation}
\begin{aligned}
    &\forall e_1 = (v^{j_1}_{t}, v^{i_1}_{t+1}, d_1), e_2 = (v^{j_2}_{t}, v^{i_2}_{t+1}, d_2) \in E_T \\
    &\qquad \implies \ d_1 = d_2
\end{aligned}
\end{equation}

%% file: appendix/extended_method_implem.tex
\section{Extended Methodology and Implementation}
\label{apdx:impl_details}

{This section expands upon the \methodName framework presented in \cref{sec:methodology_framework}. We first provide additional details on the method and report the exact prompts of the MLLM Judge and Labeler. Then, we describe the selection of the hyperparameters and the software and hardware specifications required to reproduce our results.}

\subsection{{Additional Method Details}}

\paragraph{Ensembled Mask Sampling Protocol.}
\label{apdx:mask_enasemble}
As detailed in \cref{sec:methodology_framework}, the framework extracts the sampled masks $s_j^i$ by prompting the visual decoder with a single point ($K=1$) drawn from within the spatial boundaries of each disjoint component. Relying on a single random point coordinate can introduce sampling variance. To ensure structural stability, we perform this single-point query $N=32$ times per component and ensemble the outputs to form a single, highly robust union mask $s^i_j$. During this ensembling process, we discard any individual mask proposal whose predicted quality score from the SAM decoder falls below $\tau_{\text{IoU}}=0.8$.

\paragraph{Tracker and Mask Decoder Modularity.}
While \cref{sec:methodology_framework} conceptually separates the Tracker and the Mask Decoder, our experiments instantiate both using the same foundation model (\eg, SAM 2.1 or SAM 3) to maximize inference efficiency via feature caching. However, the framework architecture is strictly modular, decoupling temporal tracking from single-frame zero-shot segmentation. This flexibility allows future deployments to pair a lightweight, tracking-only model with an independent promptable mask decoder, naturally accommodating advancements in either domain.

\paragraph{Algorithmic Bounds for State Generation.}
As formulated in \cref{sec:methodology_framework}, the State Generator enumerates all possible physical configurations by computing the partition set $\Pi(X)$ of the tracked objects or disjoint components. Because the cardinality of a partition set scales according to the $n$-th Bell number $B_n$, an unconstrained search could lead to combinatorial explosion. To ensure computational tractability in our experiments, we set a configurable parameter that limits the number of locally interacting items to $|X| \le 7$, which limits the search space at $B_7 = 877$ unique state proposals. To further optimize assembly generation, we strictly pre-filter the input space by excluding any objects whose sampled masks do not spatially overlap. {Additionally, to avoid very small noisy components in the mask to interfere, we filter each component with area smaller than 0.02\% of the image resolution, analogous to TubeletGraph~\cite{sun2026tracking}.} {The measured runtime of state generation and evaluation up to this bound is reported in Appendix~\ref{apdx:scalability}.}

Crucially, this threshold only defines the maximum degree of the state graph for a single transformation event—meaning a maximum of seven parts can assemble into one entity, or one entity can disassemble into at most seven parts simultaneously. It does \textit{not} limit the total number of independent objects the framework can continuously track in the background, which can be arbitrarily high. The value of 7 was simply chosen for our experimental setup (in \datasetName, the maximum in-degree for an assembly is 2, and the maximum out-degree for a disassembly is 3) and can easily be increased depending on available hardware and application requirements.

\subsection{{MLLM Prompts}}

\paragraph{MLLM Judge Prompts.}
The MLLM Judge serves as a visual validator to confirm whether the proposed transformation reflects actual physical changes in the scene. To provide the necessary temporal context, the model receives sparsely sampled historical frames alongside the current frame overlaid with proposed regions. A transformation is accepted into the state graph only if the model explicitly outputs ``\textsc{YES}''. The exact prompts for assembly and disassembly validation are detailed in \cref{fig:prompt_judge_assembly,fig:prompt_judge_disassembly}, respectively.

\begin{figure*}[p]
\centering
\begin{userprompt}
You are analyzing a video sequence to track objects. You are provided five images:
\begin{itemize}
    \item[-] [Image 1] Context frame (5s before).
    \item[-] [Image 2] Context frame (3s before).
    \item[-] [Image 3] Context frame (1s before).
    \item[-] [Image 4] Current frame.
    \item[-] [Image 5] Current frame with colored outlines and labels highlighting regions belonging to different tracked objects.
\end{itemize}

The outlined and labeled regions in Image 5 are: {labeled\_regions}
Task: Determine if these objects have been assembled (e.g they are physically attached/joined together) OR if they remain distinct, independent objects (e.g they are not physically attached).

You must respond ONLY with a valid JSON object using the following schema: \newline
\{\newline
  "analysis": "Provide exactly ONE short sentence comparing the context frame to the current frame to verify if physical attachment or merging occurred.",\newline
  "all\_assembled": "YES or NO. Answer YES only if confident ALL the objects have assembled into one. Answer NO if they are still separate objects or if uncertain."\newline
\}\newline
\end{userprompt}
\caption{MLLM Judge prompt for assembly validation.}
\label{fig:prompt_judge_assembly}
\end{figure*}

\begin{figure*}[p]
\centering
\begin{userprompt}
You are analyzing a video sequence to track objects. You are provided five images:
\begin{itemize}
    \item[-] [Image 1] Context frame (5s before).
    \item[-] [Image 2] Context frame (3s before).
    \item[-] [Image 3] Context frame (1s before).
    \item[-] [Image 4] Current frame.
    \item[-] [Image 5] Current frame with colored outlines and labels highlighting proposed object regions.
\end{itemize}

The outlined and labeled regions in Image 5 are: {labeled\_regions}
All the proposed object regions in Image 5 currently tracked together. 
Task: Determine if these outlined regions have separated into physically distinct objects (e.g they are no more physically attached or touching) OR if they are still a single object (e.g they are still physically attached).

You must respond ONLY with a valid JSON object using the following schema:\newline
\{\newline
  "analysis": "Provide exactly ONE short sentence comparing the context frame to the current frame to identify occlusion or breakage.",\newline
  "all\_distinct": "YES or NO. Answer YES only if confident they are distinct objects. Answer NO if they are parts of the same object."\newline
\}\newline
\end{userprompt}
\caption{MLLM Judge prompt for disassembly validation.}
\label{fig:prompt_judge_disassembly}
\end{figure*}

\paragraph{MLLM Labeler Prompts.}
Once a structural change is validated, the MLLM Labeler is queried to populate the state graph with explicit semantic labels. It assigns concise nouns to the newly formed object states and a descriptive verb phrase to the transformation edges. The exact prompts used for assembly and disassembly labeling are detailed in \cref{fig:prompt_labeler_assembly,fig:prompt_labeler_disassembly}, respectively.

\begin{figure*}[p]
\centering
\begin{userprompt}
You are analyzing a video sequence where multiple physical objects have been ASSEMBLED into a single object.

You are provided with two images:
\begin{itemize}
    \item[-] [Image 1] Context frame (5 seconds before the assembly).
    \item[-] [Image 2] Current frame right after the assembly, with a colored contour outline and letter label highlighting the assembled result.
\end{itemize}

The individual parts that were assembled together:
{before\_objects}

The assembled result (labeled in Image 2):
{labeled\_regions}

Task:
\begin{enumerate}
    \item Name the assembled result (labeled region) based on what you see in Image 2 and knowing it was formed from the parts listed above.
    \item Describe the specific assembly action (\eg "snap together", "screw on lid", "insert battery").
\end{enumerate}

You must respond ONLY with a valid JSON object using the following schema:\newline
\{\newline
  "summary": "One sentence describing what happened.",\newline
  "objects": [\{"label": "A", "name": "concise name for the assembled object"\}],\newline
  "action\_name": "concise verb phrase for the assembly action"\newline
\}
\end{userprompt}
\caption{MLLM Labeler prompt for assembly events.}
\label{fig:prompt_labeler_assembly}
\end{figure*}

\begin{figure*}[p]
\centering
\begin{userprompt}
You are analyzing a video sequence where a physical object has been DISASSEMBLED into multiple parts.

You are provided with two images:
\begin{itemize}
    \item[-] [Image 1] Context frame (5 seconds before the disassembly).
    \item[-] [Image 2] Current frame right after the disassembly, with colored contour outlines and letter labels highlighting the resulting parts.
\end{itemize}

The object that was disassembled:
{before\_objects}

The resulting parts after disassembly (labeled in Image 2):
{labeled\_regions}

Task:
\begin{enumerate}
    \item Name each resulting part (labeled region) based on what you see in Image 2 and the context from Image 1.
    \item Describe the specific disassembly action (\eg "unscrew lid", "pull apart halves", "detach battery cover").
\end{enumerate}

You must respond ONLY with a valid JSON object using the following schema:\newline
\{\newline
  "summary": "One sentence describing what happened.",\newline
  "objects": [\{"label": "A", "name": "concise part name"\},\{"label": "B", "name": "concise part name"\}],\newline
  "action\_name": "concise verb phrase for the disassembly action"\newline
\}
\end{userprompt}
\caption{MLLM Labeler prompt for disassembly events.}
\label{fig:prompt_labeler_disassembly}
\end{figure*}

{
\subsection{{Hyperparameter Selection}}
\label{apdx:hyperparams}

{With the exception of the state selection threshold $\tau$ of \methodName \textit{w/o Judge}, which is selected by cross-validation (\cref{fig:tau_cv}), no hyperparameter was tuned on VISTA;} we selected them following standard or conservative values.

\begin{itemize}
  \item $\tau$: threshold used in the state selection (\cref{sec:state_selec}). In the default configuration of \methodName (with MLLM Judge) we use $\tau = 0.5$. This comes from the interpretation of the overlap scores as proxy measure for confidence in the proposed state. A state that has overlap score below 0.5 is considered spurious, therefore not considered in the final selection. When removing the MLLM Judge (\textit{\methodName w/o Judge} configuration), the threshold is increased to $\tau = 0.8$ to compensate for the lack of reasoning from the Judge{, as selected by the cross-validation protocol described below}. 
  
  \item $K$: number of points sampled to decode object masks (\cref{sec:comp_extr_samp}). In \cref{tab:sampled_points_ablation} we ablated the choice of use $K=1$. This is required to avoid a change in behavior from SAM's mask decoder. When using $K\ge2$ we remove all the ambiguity in the mask decoding process. This is a fundamental flaw because SAM will start decoding mostly the specific component we prompt. In contrast, when $K=1$, SAM is forced to decode multiple masks and score and select the most likely, solving occlusions, and providing the informative signal needed to evaluate quickly the proposed states. 

  \item $N$: number of masks sampled for component (Appendix~\ref{apdx:mask_enasemble}). This is set by default to $N=32$, a fairly large value to reduce the variance of the sampling process. We observed that reducing $N$ to 8, 4 and 1 has little effect on the results but for correctness we want to provide a more robust sampled mask, specifically to avoid very unlucky cases due to the random sampling of the decoding points.

  \item $\tau_{IoU}$: threshold used to ensemble decoded masks (Appendix~\ref{apdx:mask_enasemble}). This is set by default to $\tau_{IoU}=0.8$ reflecting the convention of SAM to filter high quality masks.

\end{itemize}
}

\input{figures/tau_cv}

{\paragraph{Threshold Selection without MLLM Judge.}
\datasetName does not provide a separate labelled validation split. Therefore, to select $\tau$ for \methodName \textit{w/o Judge} without fitting the evaluation sequences, we adopt a repeated $k$-fold cross-validation over the $50$ sequences of \datasetName. We sweep $\tau \in \{0.5, 0.6, 0.7, 0.8, 0.9\}$ (seed 0) and, for each of $10$ repetitions with a different random split, we partition the sequences into $10$ folds. In each fold, we select the threshold that maximizes the mean $\mathcal{T}_{F_1}$ on the $45$ training sequences and evaluate it on the $5$ held-out ones, for a total of $100$ selections. Crucially, we do not tune $\tau$ for a specific tracker: the selection criterion is the $\mathcal{T}_{F_1}$ averaged over SAM 2.1 and SAM 3, yielding a single threshold shared by both trackers.

As shown in \cref{fig:tau_cv} (right), $\tau = 0.8$ is selected in $97$ out of $100$ folds, and $\tau = 0.9$ in the remaining $3$. The sensitivity analysis in \cref{fig:tau_cv} (left) explains this choice: increasing $\tau$ from $0.5$ to $0.8$ improves $\mathcal{T}_{F_1}$ by $+13.8$ and $+9.9$ points for SAM 2.1 and SAM 3 respectively. Beyond $\tau = 0.8$, SAM 2.1 saturates ($+0.8$ $\mathcal{T}_{F_1}$ points) while SAM 3 drops by $-4.0$ points, since valid transformations start to be rejected. The selected $\tau = 0.8$ is then frozen and used in all the \methodName \textit{w/o Judge} experiments. This confirms that a single threshold transfers across trackers without any tracker-specific tuning.}

\subsection{{Software and} Hardware Specifications}
All experiments, including pure tracking and MLLM inference, were conducted using two NVIDIA A100 64GB GPUs. {All methods are implemented and evaluated using Python 3.12.13 and PyTorch 2.7.0 (CUDA 12.6). All open-weight MLLMs and LLMs are self-hosted using vLLM 0.23.0. To improve reproducibility, we set the decoding temperature to $0$ for all LLMs and MLLMs whenever the model provider allows it. The default seed is 0 for Python, PyTorch and Numpy random number generators. Multi-seed experiments use seeds $\{0, 1, 2\}$}.

%% file: figures/tau_cv.tex
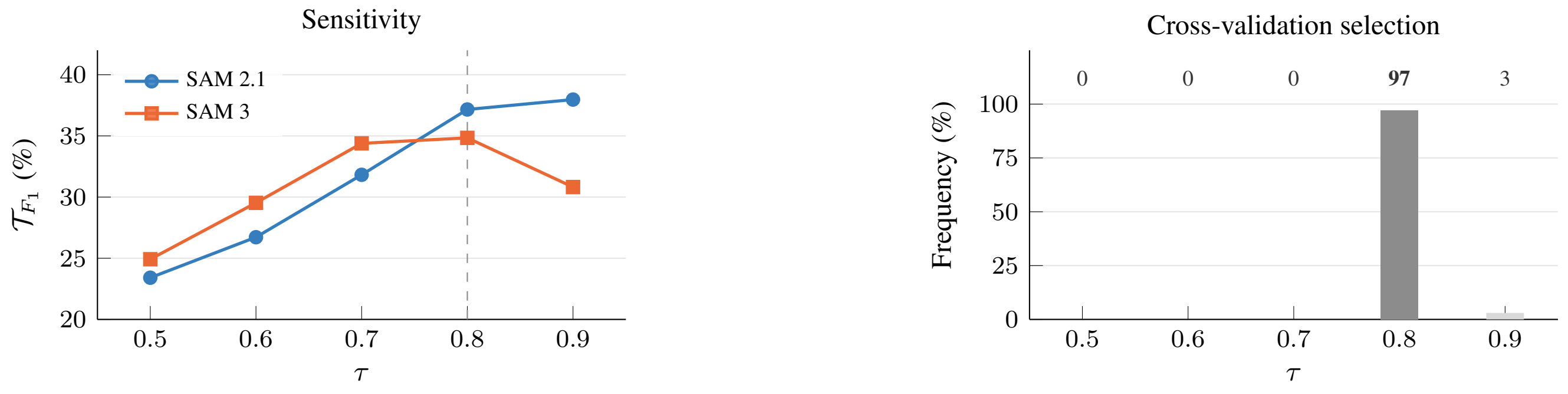
\begin{figure*}[t]

\centering
\pgfplotsset{
    taucv/.style={
        width=0.43\linewidth, height=4.6cm,
        xmin=0.45, xmax=0.95, xtick={0.5,0.6,...,0.9},
        xticklabel style={/pgf/number format/.cd, fixed, fixed zerofill, precision=1},
        xlabel={$\tau$},
        axis line style={black}, tick style={black},
        tick label style={font=\footnotesize, black}, label style={font=\small, black},
        title style={font=\small, black, yshift=-1ex},
        axis x line*=bottom, axis y line*=left,
    },
    tausens/.style={
        taucv, title={Sensitivity},
        ymin=20, ymax=42, ytick={20,25,...,40}, ylabel={$\mathcal{T}_{F_1}$ (\%)},
        grid=major, major grid style={black!10}, xmajorgrids=false,
        every axis plot/.append style={line width=1pt, mark size=1.8pt},
        legend style={font=\scriptsize, text=black, draw=none, fill=white, fill opacity=0.85, text opacity=1,
            at={(0.03,0.97)}, anchor=north west, legend cell align=left},
    },
    tauvote/.style={
        taucv, title={Cross-validation selection},
        ymin=0, ymax=125, ytick={0,25,50,75,100}, ylabel={Frequency (\%)},
        grid=major, major grid style={black!10}, xmajorgrids=false,
    },
    taubar/.style={ybar, bar width=12pt, bar shift=0pt, fill=black!15, draw=none},
    taubarsel/.style={ybar, bar width=12pt, bar shift=0pt, fill=black!45, draw=none},
    tauvals/.style={
        mark=none, draw=none, point meta=explicit symbolic,
        nodes near coords, every node near coord/.style={font=\scriptsize, black!80, anchor=center},
    },
}
\begin{tikzpicture}
\begin{axis}[tausens]
\draw[dashed, black!45] (axis cs:0.8,20) -- (axis cs:0.8,42);
\addplot[cvprblue, mark=*] coordinates {(0.5,23.41) (0.6,26.72) (0.7,31.82) (0.8,37.16) (0.9,37.97)};
\addlegendentry{SAM 2.1}
\addplot[tipoDis, mark=square*] coordinates {(0.5,24.92) (0.6,29.53) (0.7,34.39) (0.8,34.84) (0.9,30.82)};
\addlegendentry{SAM 3}
\end{axis}
\end{tikzpicture}
\hfill
\begin{tikzpicture}
\begin{axis}[tauvote]
\addplot[taubar] coordinates {(0.5,0) (0.6,0) (0.7,0) (0.9,3)};
\addplot[taubarsel] coordinates {(0.8,97)};
\addplot[tauvals] coordinates {(0.5,112) [0] (0.6,112) [0] (0.7,112) [0] (0.8,112) [\textbf{97}] (0.9,112) [3]};
\end{axis}
\end{tikzpicture}
\caption{{\textbf{State selection threshold of \methodName \textit{w/o Judge}.} \textbf{(left)} Transformation tracking quality $\mathcal{T}_{F_1}$ on \datasetName as a function of the state selection threshold $\tau$ for SAM 2.1 and SAM 3. \textbf{(right)} Frequency with which each threshold is selected across the $100$ folds of a $10\times10$-fold cross-validation, using a single threshold shared by both trackers. The selected value $\tau=0.8$ is highlighted.}}
\label{fig:tau_cv}
\end{figure*}

%% file: appendix/dataset.tex
\section{The \datasetName Benchmark}
\label{apdx:dataset}

This section expands upon the \datasetName benchmark introduced in \cref{sec:dataset}. {First, we describe the composition of the dataset. Then,} we detail the custom annotation pipeline and expert protocol utilized to guarantee high-fidelity ground-truth state graphs for the \taskFullName (\taskShortName) task.
{Finally,} we formalize the mathematical definitions for our proposed tracking and state-graph evaluation metrics.

{
\subsection{Dataset Composition}
\label{apdx:composition}
\datasetName comprises 50 video sequences collected from three sources: 17 videos from Assembly101~\cite{sener2022assembly101}, 8 videos from IKEA ASM~\cite{ben2021ikea}, and 25 custom recordings. The sequences cover a diverse set of object categories, including generic household objects (\eg, USB sticks, perfume boxes, moka pots, mixers, remote controls, and digital cameras), IKEA furniture (\eg, tables and drawers), and toy vehicles from Assembly101 (\eg, bulldozers, garbage trucks, and excavators), for a total of 218 unique object and part names. The annotated transformations are described by 43 distinct action descriptions, such as ``disconnect'', ``insert'', ``mount'', ``pop'', ``pull out'', ``put'', ``remove'', ``screw'', ``separate'', and ``unscrew''. All annotations have been manually checked and validated by an expert annotator to ensure the temporal and spatial validity of each transformation.
}

\subsection{Annotation Protocol and Interface}
\label{apdx:annot}
\input{figures/anno_tool}

To construct the \datasetName dataset, we designed a custom annotation interface (\cref{fig:anno_tool}) that couples automated object tracking with expert human verification. The ground-truth annotations require two modalities: dense spatial masks and discrete state transformations. Masks are generated by an annotator assisted by a Promptable Visual Segmentation tracker (SAM 3~\cite{carion2026sam}), while the transformations are manually logged to construct the ground-truth state graph. {Each sequence was annotated by one annotator and independently reviewed by a second; disagreements on transformation timing, structure, or labels were discussed and resolved until both agreed.}

\paragraph{Registering and Tracking Entities.} 
The annotator initiates the process by defining a new object identity and clicking the target within the interactive video player. They then trigger the visual tracker for that specific item (\cref{fig:anno_tool}.1, \cref{fig:anno_tool}.2). To ensure rigorous structural definitions, the human expert is explicitly tasked with identifying the precise moment a transformation is \textit{visually completed}. Annotators are instructed to log a state change only when the physical action is entirely finalized in the visual record. Upon this exact frame of completion, the expert terminates the prior tracks and registers the resulting post-transformation item(s) under entirely new identities.

\paragraph{Logging Transformations.} 
Once the pre- and post-transformation tracks are securely registered, the expert explicitly links them. By opening the dialog interface (\cref{fig:anno_tool}.3), the annotator connects the corresponding identities and provides a concise textual description of the transformation action. This registers new edges in the state graph. The recorded transformations remain continuously editable in the side panel (\cref{fig:anno_tool}.4), allowing the expert to refine semantic labels until the entire video sequence is exhaustively and accurately modeled.

\subsection{Evaluation Metrics}
\label{apdx:eval_metrics}

\paragraph{Mathematical Preliminaries.}
To formalize the evaluation metrics, we first define the ground-truth structures. The set of all masks at frame $t$ is denoted as $\mathcal{M}_t = \{ m_t^i \mid \exists i, n : (t, i, m_t^i, n) \in V \}$, and the complete set of valid frame-wise masks across the video is $\mathcal{M} = \{ \mathcal{M}_t \mid \mathcal{M}_t \neq \emptyset \}$. The unique objects present in the state graph are $\mathcal{O} = \{ i \mid \exists t, m_t^i, n : (t, i, m_t^i, n) \in V \}$.

A transformation $k$ within the state graph is defined by its timestamp $t_k$, its textual description $d_k$ (excluding trivial spatial motion where $d=\textsc{none}$), and the identities of the objects involved before and after the event. We denote the set of input object IDs as $T_k^\text{(pre)}$ and the set of resulting output object IDs as $T_k^\text{(post)}$. We define the chronologically ordered set of all transformations in the ground-truth graph as:
\begin{equation}
\begin{aligned}
    &T = \{ (k, t_k, T_k^\text{(pre)}, T_k^\text{(post)}, d_k) \mid{} \\
    &\qquad T_k^\text{(pre)} \neq \emptyset \land T_k^\text{(post)} \neq \emptyset \}
\end{aligned}
\end{equation}
The temporal ordering is foundational for the transformation matching protocol we will discuss below.

\paragraph{Tracking Quality.}
Following VOST~\cite{tokmakov2023breaking}, we compute tracking quality $\mathcal{J}$ as the frame-wise Intersection-over-Union (IoU) between the ground-truth mask set $\mathcal{M}$ and the predicted mask set $\mathcal{\hat M}$:
\begin{equation}
    \mathcal{J}(\mathcal{M}, \mathcal{\hat{M}}) = \frac{1}{|\mathcal{M}|}\sum_{\mathcal{M}_t \in \mathcal{M}} \text{IoU}\left(\bigcup_{m_t^i \in \mathcal{M}_t} m_t^i, \bigcup_{\hat m_t^i \in \mathcal{\hat M}_t} \hat m_t^i\right)
\end{equation}
To measure tracking persistence, $\mathcal{J}_{tr}$ computes this same metric exclusively over the final 25\% of the video frames. Unlike VOST, we intentionally compute $\mathcal{J}$ globally per frame rather than on a strictly per-object basis. Enforcing a per-object mapping would entangle $\mathcal{J}$ with the quality of transformation detection, defeating its primary purpose. By evaluating spatial overlap globally, we fully decouple pure vision tracking capabilities from state-graph generation accuracy. For instance, if a baseline method (\eg, Vanilla SAM) fails to detect any transformations but perfectly tracks the spatial footprints of all disjoint parts, its $\mathcal{J}$ and $\mathcal{J}_{tr}$ scores remain unpenalized, isolating its tracking performance.

\begin{figure*}[tp]
\centering
\begin{systemprompt}
You are a highly intelligent assistant that can analyze actions and resulting objects in text.
\end{systemprompt}
\vspace{0.5em}
\begin{userprompt}
Given the object description '[GT\_OBJECT]', is '[PRED\_OBJECT]' similar to it? Please rate from -1 to 1, where -1 means completely unrelated, 0 means ambiguous, and 1 means '[PRED\_OBJECT]' is similar. Over- or under-specified descriptions should still be considered as +1. Please answer with a single integer.
\end{userprompt}
\caption{LLM evaluator prompt for assessing object naming semantic accuracy ($\mathcal{S}_O$).}
\label{fig:prompt_eval_object}
\end{figure*}

\begin{figure*}[tp]
\centering
\begin{systemprompt}
You are a highly intelligent assistant that can analyze actions in text.
\end{systemprompt}
\vspace{0.5em}
\begin{userprompt}
Given a particular action description of '[GT\_ACTION]', is '[PRED\_ACTION]' similar to the verbs in this action? Please rate from -1 to 1, where -1 means completely unrelated, 0 means ambiguous, and 1 means '[PRED\_ACTION]' captures the meaning of '[GT\_ACTION]' or is directly in it. Brief/general descriptions should still be considered as +1. Please answer with a single integer.
\end{userprompt}
\caption{LLM evaluator prompt for assessing action description semantic accuracy ($\mathcal{S}_A$).}
\label{fig:prompt_eval_action}
\end{figure*}

\paragraph{Transformation Tracking ($\mathcal{T}_{F_1}$).}
To evaluate the predicted state graph, we establish a bipartite matching between the predicted transformations $\hat{T}$ and the ground-truth transformations $T$. This mapping allows us to compute precision ($\mathcal{T}_p$), recall ($\mathcal{T}_r$), and the macro-averaged F1-score ($\mathcal{T}_{F_1}$):
\begin{equation}
\begin{aligned}
    \mathcal{T}_p &= \frac{TP}{TP + FP} \\
    \mathcal{T}_r &= \frac{TP}{TP + FN} \\
    \mathcal{T}_{F_1} &= \frac{2\cdot TP}{2 \cdot TP + FN + FP}
\end{aligned}
\end{equation}
where True Positives ($TP$) represent correctly matched transformations, False Positives ($FP$) are hallucinated transformations in the prediction, and False Negatives ($FN$) are undetected ground-truth transformations. All metrics are computed at the sequence level and macro-averaged.

The bipartite matching is resolved using a modified Jonker-Volgenant algorithm~\cite{crouse2016implementing}, governed by a strict cost matrix $C_{p,g}$ between a predicted transformation $p$ and a ground-truth transformation $g$:
\begin{equation}
    C_{p,g} = 
    \begin{cases}
        +\infty  & \text{if } |\hat T^{(\text{pre})}_p| \ne |T^{(\text{pre})}_g| \lor{} \\
                 & \phantom{\text{if }} |\hat T^{(\text{post})}_p| \ne |T^{(\text{post})}_g| \\
        +\infty  & \text{if } \hat t_p \le t_{g-1} \lor \hat t_p \ge t_{g+1} \\
        +\infty  & \text{if } \mathcal{T}_{\text{check}}(p, g) = 0 \\ 
        |\hat t_p - t_g| & \text{otherwise}
    \end{cases}
\end{equation}
This cost matrix enforces three physical realities:
\begin{enumerate}[leftmargin=*, nosep]
    \item \textbf{Object Degree Matching:} The first condition assigns infinite cost if the transformations have a different number of input or output objects, ensuring assemblies are never matched to disassemblies.
    \item \textbf{Temporal Boundaries:} The second condition prevents a prediction from matching a ground-truth event $g$ if it occurs outside the temporal window defined by the preceding ($t_{g-1}$) and succeeding ($t_{g+1}$) ground-truth transformations. In case $g$ is the first or the last ground-truth transformation, we manually define $t_{g-1}=0$ and $t_{g+1}=+\infty$, respectively. Moreover, in the edge case that $t_g = t_{g-1}$ or $t_g = t_{g+1}$, we recursively take the previous ($g-2$) or next ($g+2$) transformation until the frame time is different to avoid assigning a wrong $+\infty$ penalty. {This is the widest admissible window since $(t_{g-1}, t_{g+1})$ is the largest interval around $t_g$ that contains no other ground-truth transformation. Hence, within this window the ground-truth state graph changes only through $g$ itself. Extending it past either bound would compare $p$ against objects that do not exist yet (\eg, inputs of $g$ produced by $g-1$) or no longer exist (\eg, outputs of $g$ consumed by $g+1$), leaving $\mathcal{T}_{\text{check}}$ ill-defined.}
    \item \textbf{Spatial Consistency:} The third condition uses a boolean function $\mathcal{T}_{\text{check}}(p, g)$ which returns $1$ only if every object in the pre- and post-transformation sets achieves an IoU $ > 0.5$ between prediction and ground truth.
\end{enumerate}

Crucially, the spatial consistency check $\mathcal{T}_{\text{check}}$ is non-trivial and must account for two complex tracking dynamics. First, to handle arbitrary tracking ID assignments, an internal bipartite matching is performed on the object IDs to optimally align the prediction with the ground truth. Second, a predicted transformation $p$ and ground-truth $g$ may be temporally shifted, during which a prediction might be hallucinated, generating false-positive intermediate states. To mitigate this alignment problem, the IoU matching is evaluated at frame $\max(\hat{t}_p, t_g)$, and we verify that the matched objects correctly trace their structural ancestry back to the original $\hat{T}^{(\text{pre})}_p$ or $\hat{T}^{(\text{post})}_p$ sets, depending on the selected frame. 

{\paragraph{Comparison with TAS Metrics.}
\label{apdx:tas_metrics}
Existing TAS metrics~\cite{sun2026tracking} are constrained by the VOST-TAS annotations, which contain only the resulting parts of each transformation{, annotated at a single frame at the end of its window,} and no state graph. Consequently, TAS metrics match transformations on their outputs alone and cannot verify which objects take part in them, making them insufficient to evaluate \taskShortName. For disassemblies, the matching is incomplete: if the prediction splits an object $A$ into $\{B, C, D\}$ while the ground truth is $A \rightarrow \{B, C\}$, the transformation is still matched and $D$ is left unaccounted for. For assemblies, the issue is more severe: since only the resulting object is checked, a prediction $\{A, C\} \rightarrow E$ is matched to the ground-truth $\{A, B\} \rightarrow E$ even though the wrong parts were assembled. {Moreover, the spatial check of $\mathcal{H}_{ST}$ is performed on the tracked objects at the last frame of the ground-truth transformation, rather than on the objects produced by the matched prediction. Hence, a prediction with the wrong number of parts is credited whenever later, unmatched transformations repair the tracks before that frame.} In contrast, $\mathcal{T}_{F_1}$ leverages the full state graphs of \datasetName: the object degree condition rejects the first case, and $\mathcal{T}_{\text{check}}$ rejects the second, requiring every pre- and post-transformation object to match.}

\input{tables/id_switches}

{\paragraph{Identity Switches.}
\label{apdx:id_switches}
Since $\mathcal{J}$ is computed on union-level masks, identity switches must be handled by the state-graph metrics. We distinguish two types of identity switches. First, a tracked object may drift to an untracked object. In this case, the predicted mask no longer overlaps with its ground-truth object, which is directly penalized by $\mathcal{J}$ and $\mathcal{J}_{tr}$, and by $\mathcal{T}_{F_1}$ whenever the object is involved in a subsequent transformation, since $\mathcal{T}_{\text{check}}$ requires an IoU $ > 0.5$ for every involved object. Second, two tracked objects may swap their identities. Since the union of their masks is unchanged, this event is not captured by $\mathcal{J}$. It is penalized by $\mathcal{T}_{F_1}$ only if one of the two objects is subsequently involved in a transformation and the objects do not swap back before it occurs. Conversely, if both objects are involved in the same transformation, the internal bipartite matching of $\mathcal{T}_{\text{check}}$ absorbs the permutation and the swap is not penalized.

This second type is a very specific event, as it requires two tracked objects to exchange their entire spatial footprints, which is rare for trackers with per-object memory such as SAM 2.1 and SAM 3. To verify that its impact on our evaluation is negligible, we explicitly count this type of identity swaps between tracked objects. At every frame, we match the predicted tracks to the ground-truth objects through a bipartite matching on mask IoU, retaining only pairs with IoU $> 0.5$, as in $\mathcal{T}_{\text{check}}$. Each track $i$ stores its current match $c_i$, \ie, the ground-truth object it was last matched to, and its previous match $p_i$, \ie, the ground-truth object it held before $c_i$. A swap between tracks $i$ and $j$ is counted whenever one of them changes its match such that $p_i = c_j$ and $p_j = c_i$, with both tracks matched at that frame. Since the two tracks need neither to change their match at the same frame nor to be matched at every frame, this also captures swaps that occur progressively or while the tracks are temporarily unmatched. Transformations assign new identities to their output objects, both in the predictions and in the ground truth, so the identity changes they cause are never counted as swaps. Each exchange is counted as a separate event, including a subsequent swap back, and we report the total over all sequences. As shown in \cref{tab:id_switches}, no identity swap occurs for any \methodName configuration.}

\paragraph{Transformation Semantics.}
Following TubeletGraph~\cite{sun2026tracking}, we utilize an LLM evaluator ({Gemma 4 31B~\cite{gemma42026}} in our case) to assess the semantic similarity between predicted and ground-truth text labels for all correctly matched transformations ($TP$). 
To evaluate object naming accuracy ($\mathcal{S}_O$), the model is prompted with the instructions detailed in \cref{fig:prompt_eval_object}. 
To evaluate action description accuracy ($\mathcal{S}_A$), the model is prompted with the instructions detailed in \cref{fig:prompt_eval_action}.
A label is deemed correct only if the model returns $1$. 
The overall semantic quality $\mathcal{S}$ per sequence is calculated as the harmonic mean of $\mathcal{S}_O$ and $\mathcal{S}_A$ and then macro-averaged across sequences.

\input{tables/human_agreement}

{To select the LLM evaluator, we measure its agreement with human judgments. Six human subjects each evaluate 92 unique (ground-truth, predicted) label pairs, randomly sampled from the correctly matched transformations of \datasetName and covering both object names and action descriptions. For each pair, subjects provide a binary judgment on whether the predicted label is semantically similar to the ground-truth one, consistently with the binarized decision of the LLM evaluator. The agreement is computed as the percentage of pairs for which the human and the LLM decisions coincide, using fixed prompts for all models. As shown in \cref{tab:human_agreement}, Gemma 4 31B achieves the highest agreement with human judgments ($88.3\%$ vs $81.1\%$ for GPT-5.4 mini), while also being an open-weight model. We therefore adopt it as the LLM evaluator for all experiments.}

\paragraph{Overall Score ($\mathcal{H}_{F_1}$).}
The overall \taskShortName quality is measured by computing the F1-score as in $\mathcal{T}_{F_1}$, but strictly filtering the True Positives ($TP$) to only include transformations that achieve perfect semantic accuracy (score of $1$ for both object and action naming).

%% file: figures/anno_tool.tex
\begin{figure*}[p]
    \centering
    \includegraphics[page=1, width=\textwidth,height=0.8\textheight,keepaspectratio]{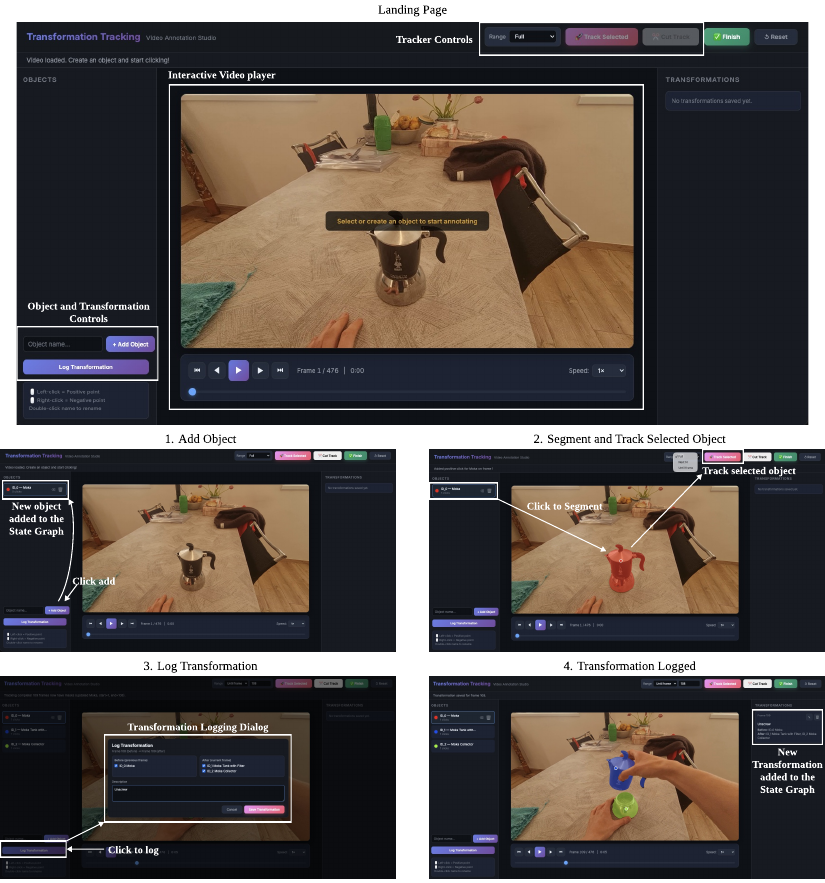}
    \caption{\textbf{Overview of the \datasetName Annotation Tool.} Our custom interface couples automated object tracking with expert human verification to construct high-fidelity state graphs. \textbf{(Top)} The landing page features an interactive video player, central tracker controls, and lateral panels for object and graph management. \textbf{(1) Add Object:} The annotator registers a new distinct entity ID in the system. \textbf{(2) Segment and Track:} By clicking on the object within the video frame, the annotator prompts the segmentation model to extract the spatial mask and initiates temporal tracking. \textbf{(3) Log Transformation:} At the exact frame of a physical state change, the expert opens a dialog box to explicitly link the pre- and post-transformation object tracks and assign a semantic action label (\eg, ``Unscrew''). \textbf{(4) Transformation Logged:} The validated event is immediately committed to the state graph and displayed in the right-hand panel for continuous review.}
\label{fig:anno_tool}
\end{figure*}

%% file: tables/id_switches.tex
\begin{table}[t]

\centering
\footnotesize
\caption{\textbf{Identity swaps on \datasetName.} Number of identity swaps between tracked objects.}
\label{tab:id_switches}

\begin{tabular*}{\linewidth}{@{\hspace{6pt}\extracolsep{\fill}}llc@{\hspace{6pt}}}
\toprule
Tracker & Method & Swaps $\downarrow$ \\
\midrule
\multirow{2}{*}{SAM 2.1}
 & \methodName \textit{w/o Judge} & 0 \\
 & \methodName & 0 \\
\midrule
\multirow{2}{*}{SAM 3}
 & \methodName \textit{w/o Judge} & 0 \\
 & \methodName & 0 \\
\bottomrule
\end{tabular*}
\end{table}

%% file: tables/human_agreement.tex
\begin{table}[t]

\centering
\footnotesize
\caption{\textbf{Human agreement of the LLM evaluator.} Agreement between human judgments and LLM evaluator on the semantic similarity of object names and action descriptions. Values are mean $\pm$ standard deviation across subjects. Best result is in \textbf{bold}.}
\label{tab:human_agreement}

\begin{tabular*}{\linewidth}{@{\hspace{6pt}\extracolsep{\fill}}lc@{\hspace{6pt}}}
\toprule
LLM Evaluator & Human Agreement (\%) $\uparrow$ \\
\midrule
GPT-5.4 mini \cite{gpt542026} & 81.1 $\pm$ 4.2 \\
Gemma 4 31B \cite{gemma42026} & \textbf{88.3 $\pm$ 3.3} \\
\bottomrule
\end{tabular*}
\end{table}

%% file: appendix/additional_quantitative.tex
\section{Additional Quantitative Analysis}
\label{apdx:extra_quant}

This section provides expanded ablation studies and a variability analysis to characterize the robustness of \methodName.

\subsection{Additional Ablations}

\input{tables/ext_ablation}

\paragraph{State Selection Thresholds.} 
In the \methodName framework, the MLLM Judge provides the high-level reasoning required to validate states at lower confidence intervals. As shown in \cref{tab:no_vlm_threshold_ablation}, \methodName remains highly effective even in its standalone configuration (\textit{w/o Judge}). However, to compensate for the absence of the MLLM's spatial reasoning, we increase the state selection threshold from $\tau=0.5$ to $\tau=0.8${, selected by cross-validation as detailed in Appendix~\ref{apdx:hyperparams}}. This adjustment yields significant improvements in transformation tracking quality, specifically $+15.2$ and $+9.7$ $\mathcal{T}_{F_1}$ points for SAM 2.1 and SAM 3 respectively. This demonstrates that while the MLLM Judge is the optimal configuration, the mask overlap scoring component is sufficiently robust to maintain accuracy through stricter filtering.

\paragraph{MLLM Judge Selection.} 
The choice of MLLM Judge is a primary factor in the system's inference speed and state validation accuracy. In \cref{tab:vlm_ablations_supp}, we compare four open-weight MLLMs. To prioritize reproducibility and minimize latency, we exclude closed-source models (\eg, GPT, Claude). Qwen 3.5 35B delivers the best all-around performance. Its Mixture of Experts (MoE) architecture allows it to achieve the highest $\mathcal{T}_{F_1}$ scores while maintaining superior frame rates compared to dense models like Gemma 4 31B.

\input{tables/tracker_size}

{\paragraph{Tracker Size.}
\label{apdx:tracker_size}
Since \methodName evaluates structural changes directly from the tracker outputs, we analyze how transformation quality scales with the size of the base tracker. In \cref{tab:tracker_size}, we evaluate SAM 2.1 in its Tiny, Base Plus, and Large (default) configurations. Transformation tracking quality consistently improves with larger trackers, especially without the MLLM Judge: $\mathcal{T}_{F_1}$ increases from $28.7$ (Tiny) to $39.5$ (Large), yielding $+10.8$ points. When the MLLM Judge is active, this gap reduces to $+6.3$ $\mathcal{T}_{F_1}$ points, as visual verification compensates for the noisier state proposals produced by smaller trackers. This confirms that \methodName scales with stronger foundation models, while the MLLM Judge provides robustness in computationally constrained settings.}

\input{tables/ext_results}

\subsection{Variability Analysis}

\input{tables/robustness}

To assess the stability of \methodName, we report the mean and standard deviation across three independent experiments using different random seeds $\{0, 1, 2\}$ {on \datasetName} in \cref{tab:tuvos_ext_main}. 

We observe minimal variability in pure tracking and transformation tracking scores. The highest observed variation is $\pm4.9$ points in the semantic score $\mathcal{S}$, which is expected given the inherent stochasticity in MLLM generative labeling. Notably, this variability is confined to the descriptive naming of objects and actions and does not impact the framework's ability to detect and track physical transformations.

{\subsection{Scalability Analysis}
\label{apdx:scalability}

\input{figures/scalability}
\input{tables/last_frame}

We profile the runtime of each \methodName module using SAM 3 as the number of components and objects in the scene grows. We omit the MLLM Judge, since it is not queried at every frame thanks to the state evaluation filtering, and its processing time is independent of the number of objects and parts. Component Extraction requires $2.5$\,ms per tracked object, and the Point Sampler scales linearly with the total number of extracted components, requiring approximately $18$\,ms per sampled component ($22$\,ms for a single component and $128$\,ms for seven).

As shown in \cref{fig:scalability}, the cost of state generation and evaluation follows the cardinality of the partition set $\Pi(X)$ (Appendix~\ref{apdx:impl_details}). Disassembly runtime increases from below $0.1$\,ms for a single component to $170$\,ms for seven components, the maximum allowed in our experiments. Assembly remains cheaper, increasing from $0.2$\,ms to $105$\,ms for seven overlapping objects, since non-overlapping objects are pre-filtered and do not contribute to the search space. However, the worst case is almost never reached in practice. On \datasetName, $95.1\%$ of the disassembly evaluations involve a single component and only $0.3\%$ involve three or more, while $76.3\%$ of the assembly evaluations involve at most two overlapping objects and fewer than $0.1\%$ reach the bound of seven (bars in \cref{fig:scalability}).

Our implementation is built purely in Python and PyTorch, and a native C++ or multi-threaded implementation would further reduce the runtime. Even in this unoptimized form, \methodName processes videos significantly faster than TubeletGraph (\cref{tab:tuvos_main}). This confirms that, when multiple objects are present, \methodName scales comparably to the underlying tracker.}

\subsection{Robustness to Tracker Noise}
\label{apdx:robustness}

Since \methodName evaluates structural changes directly from the tracker outputs, we assess its sensitivity to a degraded base tracker by corrupting the output masks of the tracker on \datasetName with random cutouts. Each frame is selected for perturbation with probability $0.2$, and three rectangular cutouts are applied at random positions, with sides uniformly sampled between 5\% and 10\% of the frame sides. The same cutouts are applied to the masks of all the tracked objects in the frame. The corruption is applied only to the tracker outputs and is not propagated to the tracker memory, while the original frames are provided to all components of the pipeline, \ie the tracker, the mask decoder, the MLLM Judge, and the MLLM Labeler. The evaluation is performed against the original ground-truth annotations. All experiments use SAM 3 as the tracker.

As shown in \cref{tab:robustness}, the corrupted masks substantially degrade pure tracking quality ($-12.2$ $\mathcal{J}$ points). As expected, the transformation quality of \methodName also decreases, with a drop of $-7.6$ $\mathcal{T}_r$, $-5.5$ $\mathcal{T}_p$, and $-6.0$ $\mathcal{T}_{F_1}$ points, but it does not collapse: \methodName still achieves $35.6$ $\mathcal{T}_{F_1}$, well above TubeletGraph with clean SAM 3 masks ($3.3$, \cref{tab:tuvos_main}). Notably, the MLLM Judge is critical in this setting: without it, the drop increases to $-12.3$ $\mathcal{T}_{F_1}$ points, suggesting that visual verification on the clean frames effectively filters the spurious state proposals induced by corrupted masks. This demonstrates that the structural reasoning of \methodName remains reliable despite significant tracker noise.

\subsection{Error Propagation Analysis}
\label{apdx:last_frame}

\input{tables/last_frame_disa}

Due to \methodName's online and streaming architecture it can be subject to some irrecoverable errors in the state graph. We want to assess whether the objects in the last frame of each video sequence are correctly identified and segmented. This proxy metric will indicate how well each method can recover to a state consistent with the observed reality. On the last frame of each video, predicted tracks and ground-truth objects are matched one-to-one with Hungarian matching on mask IoU, and a match is counted as correct if its IoU is at least $\theta$. We report micro-averaged precision, recall, and $F_1$ for $\theta = 0.5$ and $\theta = 0.75$. We compare against the base tracker alone, which keeps the initial identities throughout the video, and TubeletGraph~\cite{sun2026tracking}.

As shown in \cref{tab:last_frame}, \methodName substantially outperforms both baselines{ with either base tracker}, improving over TubeletGraph by {$+25.8$ and }$+30.1$ $F_1$ points at $\theta = 0.5${ with SAM 2.1 and SAM 3, respectively} ({$70.1$ vs $44.3$ and }$70.5$ vs $40.4$). The gain is mostly driven by precision ({$87.1$ vs $43.3$ and }$93.4$ vs $45.1$). {Since} the baselines do not model assemblies, they keep separate tracks for parts that have been merged, which results in spurious objects on the last frame. Moreover, the results of \methodName are {almost }unchanged under the stricter threshold ({$-0.8$ $F_1$ with SAM 2.1, and }$70.5$ $F_1$ at both $\theta = 0.5$ and $\theta = 0.75$ { with SAM 3}), whereas TubeletGraph drops by {$-4.6$ and }$-6.6$ points, indicating that the identities predicted by \methodName are also accurately segmented. {Notably, replacing SAM 2.1 with SAM 3 slightly improves \methodName ($70.1$ to $70.5$ $F_1$ at $\theta = 0.5$), while it degrades TubeletGraph ($44.3$ to $40.4$) due to a drop in recall ($45.3$ to $36.7$). As discussed in Appendix~\ref{apdx:tg_sam3_failure}, SAM 3 loses track of the parts after a transformation less frequently than SAM 2.1, so TubeletGraph triggers fewer tubelet searches and misses the newly formed objects.} Conversely, recall ($R$) is the critical aspect for all the methods, indicating a fundamental difficulty in tracking all the objects. We believe this is mainly driven by small objects not picked up during the transformations. We observe an example of this in {\cref{fig:quali_failures}}.

To provide the most fair comparison with TubeletGraph on the transformations it is designed to handle, we repeat the evaluation on the disassembly only events of \datasetName. As shown in \cref{tab:last_frame_disa}, TubeletGraph achieves a high precision in this setting ({$95.1$ and $89.8$ with SAM 2.1 and SAM 3, respectively}), but recovers only {$42.5\%$ and }$24.3\%$ of the objects on the last frame, as most of the parts produced by the disassemblies are not detected. In contrast, \methodName raises the recall to $56.4$ and $55.8$, more than doubling it with SAM 3. With SAM 2.1, this comes at a lower precision than TubeletGraph ($90.3$ vs $95.1$), still resulting in a gain of $+10.6$ $F_1$ points ($69.4$ vs $58.8$) at $\theta = 0.5$. With SAM 3, \methodName also improves the precision ($96.2$), resulting in a gain of $+32.3$ $F_1$ points ($70.6$ vs $38.3$). As in \cref{tab:last_frame}, the results of \methodName are {almost }unchanged at $\theta = 0.75$, whereas TubeletGraph drops by {$-5.4$ and $-7.9$} $F_1$ points. {Replacing SAM 2.1 with SAM 3 reduces the recall
 of TubeletGraph from $42.5$ to $24.3$, while the recall of \methodName is stable ($56.4$ vs $55.8$).}

%% file: tables/ext_ablation.tex
\begin{table}[t]
\centering
\caption{\textbf{Impact of State Selection Filtering.} Experiments are conducted on the VISTA dataset \textit{without} MLLM Judge averaged across three seeds. Increasing the State Selection Filtering threshold $\tau$ substantially improves state-graph transformation tracking accuracy. Values are reported as mean $\pm$ standard deviation.}
\label{tab:no_vlm_threshold_ablation}

\begin{tabular*}{\linewidth}{@{\hspace{6pt}\extracolsep{\fill}}lcccc@{\hspace{6pt}}}
\toprule
Tracker & $\tau$ & $\mathcal{T}_{r}\uparrow$ & $\mathcal{T}_{p}\uparrow$ & $\mathcal{T}_{F_1}\uparrow$ \\
\midrule
\multirow{2}{*}{SAM 2.1}
 & 0.5 & \textbf{50.4 $\pm$ 1.5} & 16.8 $\pm$ 1.0 & 21.5 $\pm$ 1.1 \\
 & 0.8 & 43.8 $\pm$ 3.6 & \textbf{35.6 $\pm$ 2.3} & \textbf{36.7 $\pm$ 3.3} \\
\midrule
\multirow{2}{*}{SAM 3}
 & 0.5 & \textbf{43.4 $\pm$ 0.7} & 21.7 $\pm$ 1.0 & 23.7 $\pm$ 1.1 \\
 & 0.8 & 37.3 $\pm$ 2.5 & \textbf{37.5 $\pm$ 2.1} & \textbf{33.4 $\pm$ 1.8} \\
\bottomrule
\end{tabular*}
\end{table}
\begin{table}[t]

\centering
\footnotesize
\caption{\textbf{MLLM Judge Ablations.} Experiments are conducted on \datasetName and averaged across three seeds. Qwen 3.5 35B is selected as the best all-around MLLM Judge considering state-graph transformation tracking accuracy and FPS. Best results per tracker are in \textbf{bold}, and second-best are \underline{underlined}.}
\label{tab:vlm_ablations_supp}

\begin{tabular*}{\linewidth}{@{\hspace{6pt}\extracolsep{\fill}}llcccc@{\hspace{6pt}}}
\toprule
Tracker & MLLM & $\mathcal{T}_{r}\uparrow$ & $\mathcal{T}_{p}\uparrow$ & $\mathcal{T}_{F_1}\uparrow$ & FPS $\uparrow$ \\
\midrule
\multirow{4}{*}{SAM 2.1}
 & Qwen 3.5 35B & \underline{48.4} & \textbf{46.7} & \textbf{44.7} & \textbf{1.8} \\
 & Gemma 4 31B & \textbf{48.5} & \underline{43.3} & \underline{41.8} & 1.4 \\
 & Qwen 3.5 9B & 43.0 & 40.8 & 39.2 & 1.4 \\
 & Gemma 4 E4B & 26.2 & 30.9 & 26.1 & \underline{1.7} \\
\midrule
\multirow{4}{*}{SAM 3}
 & Qwen 3.5 35B & \textbf{44.7} & \underline{49.0} & \underline{42.5} & \underline{2.4} \\
 & Gemma 4 31B & \underline{43.8} & 44.7 & 39.6 & 1.9 \\
 & Qwen 3.5 9B & \underline{43.8} & \textbf{51.3} & \textbf{43.4} & 2.1 \\
 & Gemma 4 E4B & 27.5 & 40.8 & 30.5 & \textbf{2.5} \\
\bottomrule
\end{tabular*}
\end{table}

%% file: tables/tracker_size.tex
\begin{table}[t]

\centering
\footnotesize
\caption{\textbf{Impact of Tracker Size.} Experiments are conducted on \datasetName using SAM 2.1 at different model sizes, with and without the MLLM Judge, using a single seed. Large is the default configuration used in all other experiments. Best results per MLLM Judge setting are in \textbf{bold}, and second-best are \underline{underlined}.}
\label{tab:tracker_size}

\begin{tabular*}{\linewidth}{@{\hspace{6pt}\extracolsep{\fill}}lcccccc@{\hspace{6pt}}}
\toprule
SAM 2.1 Size & Judge & $\mathcal{J}\uparrow$ & $\mathcal{J}_{tr}\uparrow$ & $\mathcal{T}_{r}\uparrow$ & $\mathcal{T}_{p}\uparrow$ & $\mathcal{T}_{F_1}\uparrow$ \\
\midrule
Tiny      & \multirow{3}{*}{\xmark} & 88.4 & 84.3 & 35.6 & 30.7 & 28.7 \\
Base Plus & & \underline{89.2} & \underline{85.9} & \underline{39.2} & \underline{35.3} & \underline{33.8} \\
Large     & & \textbf{90.7} & \textbf{87.8} & \textbf{47.1} & \textbf{37.2} & \textbf{39.5} \\
\midrule
Tiny      & \multirow{3}{*}{\cmark} & \underline{89.8} & \underline{86.5} & 42.9 & 38.2 & 36.7 \\
Base Plus & & 89.5 & 86.0 & \underline{44.0} & \textbf{45.3} & \underline{41.1} \\
Large     & & \textbf{90.5} & \textbf{88.4} & \textbf{47.0} & \underline{44.6} & \textbf{43.0} \\
\bottomrule
\end{tabular*}
\end{table}

%% file: tables/ext_results.tex
\begin{table*}[t]
\centering
\caption{\textbf{\methodName variability on \datasetName.} We evaluate pure tracking and state-graph generation using three different seeds. Values are reported as mean $\pm$ standard deviation. Best results per tracker are in \textbf{bold}.}
\label{tab:tuvos_ext_main}

\footnotesize
\setlength{\tabcolsep}{4pt}
\renewcommand{\arraystretch}{1.05}

\begin{tabularx}{\linewidth}{
l
l
>{\centering\arraybackslash}X
>{\centering\arraybackslash}X
>{\centering\arraybackslash}X
>{\centering\arraybackslash}X
>{\centering\arraybackslash}X
>{\centering\arraybackslash}X
}
\toprule

\multirow{2}{*}{Tracker} &
\multirow{2}{*}{Method} &
\multicolumn{2}{c}{\textsc{Tracking}} &
\multicolumn{3}{c}{\textsc{State Graph}} &
\multirow{2}{*}{FPS $\uparrow$} \\ 

\cmidrule(lr){3-4}
\cmidrule(lr){5-7}

& 
& $\mathcal{J}\uparrow$ 
& $\mathcal{J}_{tr}\uparrow$ 
& $\mathcal{S}\uparrow$ 
& $\mathcal{T}_{F_1}\uparrow$ 
& $\mathcal{H}_{F_1}\uparrow$ 
& \\ 

\midrule

\multirow{2}{*}{SAM 2.1} 
 & \methodName w/o Judge 
 & 90.4 $\pm$ 0.6 & 87.2 $\pm$ 0.9 
 & 76.8 $\pm$ 4.9 & 36.7 $\pm$ 3.3 & 19.7  $\pm$ 4.7
 & \textbf{3.0 $\pm$ 0.1} \\
 
 & \methodName 
 & \textbf{90.7 $\pm$ 0.3} & \textbf{88.6 $\pm$ 0.2} 
 & 78.9  $\pm$ 2.1 & \textbf{44.7 $\pm$ 1.6} & 27.1  $\pm$ 2.2 
 & 1.8 $\pm$ 0.0 \\ 
 
\midrule

\multirow{2}{*}{SAM 3} 
 & \methodName w/o Judge 
 & \textbf{91.5 $\pm$ 1.0} & \textbf{88.0 $\pm$ 1.4} 
 & 77.4 $\pm$ 3.4 & 33.4 $\pm$ 1.8 & 16.8 $\pm$ 0.0
 & \textbf{3.9 $\pm$ 0.1} \\
 
 & \methodName 
 & 91.4 $\pm$ 0.2 & 87.8 $\pm$ 0.5 
 & 80.2 $\pm$ 2.8 & \textbf{42.5 $\pm$ 1.5} & 28.3 $\pm$ 1.5
 & 2.4 $\pm$ 0.1 \\ 
 
\bottomrule
\end{tabularx}
\end{table*}

%% file: tables/robustness.tex
\begin{table}[t]

\centering
\footnotesize
\caption{\textbf{Robustness to tracker noise on \datasetName.} \textsc{Clean} denotes the standard setting, while \textsc{Noisy} denotes tracker output masks corrupted with random cutouts. All experiments use SAM 3 as the tracker, using a single seed. Best results per setting are in \textbf{bold}.}
\label{tab:robustness}

\setlength{\tabcolsep}{3pt}
\begin{tabular*}{\linewidth}{@{\hspace{4pt}\extracolsep{\fill}}llccccc@{\hspace{4pt}}}
\toprule
Setting & Method & $\mathcal{J}\uparrow$ & $\mathcal{J}_{tr}\uparrow$ & $\mathcal{T}_{r}\uparrow$ & $\mathcal{T}_{p}\uparrow$ & $\mathcal{T}_{F_1}\uparrow$ \\
\midrule
\multirow{2}{*}{\textsc{Clean}}
 & \methodName \textit{w/o Judge} & \textbf{92.1} & \textbf{88.6} & 39.3 & 37.7 & 34.6 \\
 & \methodName & 91.7 & 88.2 & \textbf{43.6} & \textbf{46.5} & \textbf{41.6} \\
\midrule
\multirow{2}{*}{\textsc{Noisy}}
 & \methodName \textit{w/o Judge} & \textbf{79.7} & 74.1 & 25.2 & 25.5 & 22.3 \\
 & \methodName & 79.5 & \textbf{75.2} & \textbf{36.0} & \textbf{41.0} & \textbf{35.6} \\
\bottomrule
\end{tabular*}
\end{table}

%% file: figures/scalability.tex
\begin{figure*}[t]

\centering
\pgfplotsset{
    scal/.style={
        width=0.43\linewidth, height=4.6cm,
        xmin=0.4, xmax=7.6, xtick={1,...,7},
        axis line style={black}, tick style={black},
        tick label style={font=\footnotesize, black}, label style={font=\small, black},
        title style={font=\small, black, yshift=-1ex},
    },
    scaltime/.style={
        scal, axis y line*=left, axis x line*=bottom,
        ymin=0, ymax=250, ytick={0,50,100,150,200}, ylabel={Time (ms)},
        grid=major, major grid style={black!10}, xmajorgrids=false,
        every axis plot/.append style={line width=1pt, mark size=1.8pt},
        legend style={font=\scriptsize, text=black, draw=none, fill=white, fill opacity=0.85, text opacity=1,
            at={(0.17,0.80)}, anchor=north west, legend cell align=left},
    },
    scalfreq/.style={
        scal, axis y line*=right, axis x line=none,
        ymin=0, ymax=125, ytick={0,25,50,75,100}, ylabel={Frequency (\%)},
    },
    scalbar/.style={ybar, bar width=10pt, fill=black!15, draw=none},
    scalvals/.style={
        mark=none, draw=none, point meta=explicit symbolic,
        nodes near coords, every node near coord/.style={font=\scriptsize, black!80, anchor=center},
    },
    freqlegend/.style={legend image code/.code={\fill[black!15] (0cm,-0.08cm) rectangle (0.3cm,0.12cm);}},
}
\begin{tikzpicture}
\begin{axis}[scalfreq]
\addplot[scalbar] coordinates {(1,95.12) (2,4.55) (3,0.31) (4,0.02) (5,0) (6,0) (7,0.001)};
\addplot[scalvals] coordinates {(1,115) [95.1] (2,115) [4.6] (3,115) [0.3] (4,115) [{$<$0.1}] (5,115) [0] (6,115) [0] (7,115) [{$<$0.1}]};
\end{axis}
\begin{axis}[scaltime, title={Disassembly}, xlabel={\# components}]
\addlegendimage{freqlegend}
\addlegendentry{Frequency}
\addplot[tipoDis, mark=square*] coordinates {(1,0.1) (2,0.7) (3,5) (4,13) (5,32) (6,74) (7,170)};
\addlegendentry{Runtime}
\end{axis}
\end{tikzpicture}
\hfill
\begin{tikzpicture}
\begin{axis}[scalfreq]
\addplot[scalbar] coordinates {(1,46.40) (2,29.94) (3,11.69) (4,4.76) (5,6.41) (6,0.80) (7,0.004)};
\addplot[scalvals] coordinates {(1,115) [46.4] (2,115) [29.9] (3,115) [11.7] (4,115) [4.8] (5,115) [6.4] (6,115) [0.8] (7,115) [{$<$0.1}]};
\end{axis}
\begin{axis}[scaltime, title={Assembly}, xlabel={\# overlapping objects}]
\addlegendimage{freqlegend}
\addlegendentry{Frequency}
\addplot[tipoAsm, mark=triangle*] coordinates {(1,0.2) (2,1) (3,5) (4,11) (5,25) (6,51) (7,105)};
\addlegendentry{Runtime}
\end{axis}
\end{tikzpicture}
\caption{{\textbf{State generation and evaluation runtime.} Runtime per frame using SAM 3 (lines, left axis) for \textbf{(left)} disassembly state generation and evaluation for a single object fragmented into an increasing number of components, and \textbf{(right)} assembly state generation and evaluation for an increasing number of overlapping objects. Non-overlapping objects are pre-filtered and do not contribute. Bars (right axis) report the frequency of each case among all disassembly and assembly evaluations on \datasetName.}}
\label{fig:scalability}
\end{figure*}
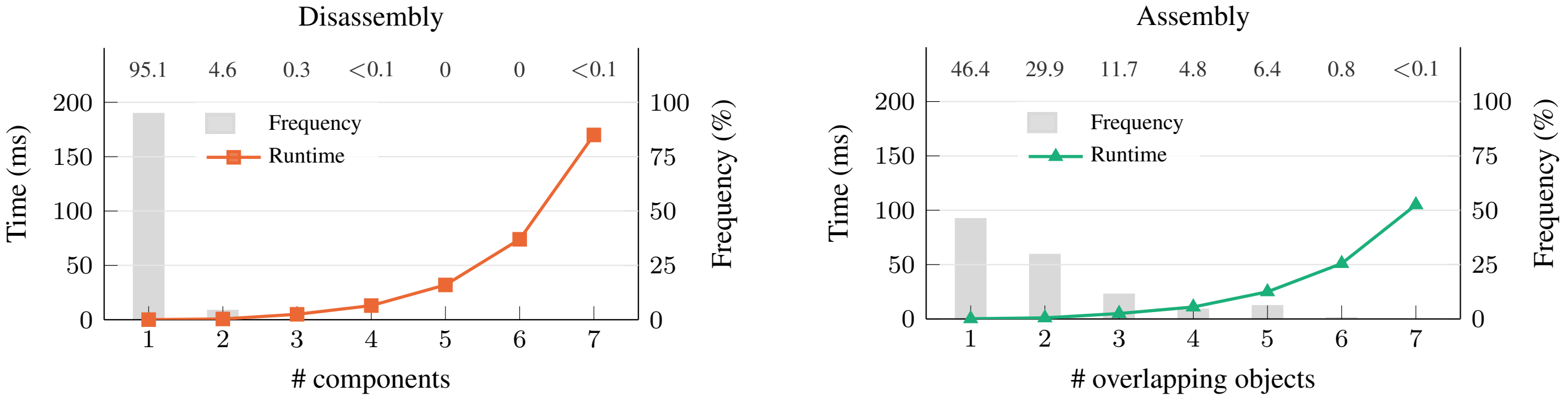

%% file: tables/last_frame.tex
\begin{table*}[t]

\centering
\footnotesize

\caption{\textbf{Last-frame object identity on \datasetName.} Predicted tracks and ground-truth objects are matched one-to-one on the last frame of each video with Hungarian matching on mask IoU if greater than $\theta$. Precision ($P$), recall ($R$) and $F_1$ are micro-averaged over all objects. {We report results using two different foundational trackers. Best results per tracker are in \textbf{bold}, and second-best are \underline{underlined}.}}
\label{tab:last_frame}

\setlength{\tabcolsep}{4pt}
\renewcommand{\arraystretch}{1.05}

\begin{tabularx}{\linewidth}{
l
l
>{\centering\arraybackslash}X
>{\centering\arraybackslash}X
>{\centering\arraybackslash}X
>{\centering\arraybackslash}X
>{\centering\arraybackslash}X
>{\centering\arraybackslash}X
}
\toprule

\multirow{2}{*}{{Tracker}} &
\multirow{2}{*}{Method} &
\multicolumn{3}{c}{$\theta = 0.5$} &
\multicolumn{3}{c}{$\theta = 0.75$} \\

\cmidrule(lr){3-5}
\cmidrule(lr){6-8}

&
& P$\uparrow$
& R$\uparrow$
& $F_1\uparrow$
& P$\uparrow$
& R$\uparrow$
& $F_1\uparrow$ \\

\midrule

\multirow{4}{*}{{SAM 2.1}}
& {\textit{Tracker only}}
& {47.8} & {35.3} & {40.6}
& {40.5} & {30.0} & {34.5} \\

& {TubeletGraph \cite{sun2026tracking}}
& {43.3} & {45.3} & {44.3}
& {38.9} & {40.7} & {39.7} \\

& {\methodName \textit{w/o Judge}}
& {\underline{83.5}} & {\underline{57.3}} & {\underline{68.0}}
& {\underline{80.6}} & {\underline{55.3}} & {\underline{65.6}} \\

& {\methodName (Ours)}
& {\textbf{87.1}} & {\textbf{58.7}} & {\textbf{70.1}}
& {\textbf{86.1}} & {\textbf{58.0}} & {\textbf{69.3}} \\

\midrule

\multirow{4}{*}{{SAM 3}}
& \textit{Tracker only}
& 43.6 & 34.0 & 38.2
& 35.9 & 28.0 & 31.5 \\

& TubeletGraph \cite{sun2026tracking}
& 45.1 & 36.7 & 40.4
& 37.7 & 30.7 & 33.8 \\

& \methodName \textit{w/o Judge}
& \underline{87.5} & \underline{56.0} & \underline{68.3}
& \underline{86.5} & \underline{55.3} & \underline{67.5} \\

& \methodName (Ours)
& \textbf{93.4} & \textbf{56.7} & \textbf{70.5}
& \textbf{93.4} & \textbf{56.7} & \textbf{70.5} \\

\bottomrule
\end{tabularx}
\end{table*}

%% file: tables/last_frame_disa.tex
\begin{table*}[t]

\centering
\footnotesize

\caption{\textbf{Last-frame object identity on the disassembly sequences of \datasetName.} Predicted tracks and ground-truth objects are matched one-to-one on the last frame of each video with Hungarian matching on mask IoU if greater than $\theta$. Precision ($P$), recall ($R$) and $F_1$ are micro-averaged over all objects. {We report results using two different foundational trackers. Best results per tracker are in \textbf{bold}, and second-best are \underline{underlined}.}}
\label{tab:last_frame_disa}

\setlength{\tabcolsep}{4pt}
\renewcommand{\arraystretch}{1.05}

\begin{tabularx}{\linewidth}{
l
l
>{\centering\arraybackslash}X
>{\centering\arraybackslash}X
>{\centering\arraybackslash}X
>{\centering\arraybackslash}X
>{\centering\arraybackslash}X
>{\centering\arraybackslash}X
}
\toprule

\multirow{2}{*}{Tracker} &
\multirow{2}{*}{Method} &
\multicolumn{3}{c}{$\theta = 0.5$} &
\multicolumn{3}{c}{$\theta = 0.75$} \\

\cmidrule(lr){3-5}
\cmidrule(lr){6-8}

&
& $P$$\uparrow$
& $R$$\uparrow$
& $F_1\uparrow$
& $P$$\uparrow$
& $R$$\uparrow$
& $F_1\uparrow$ \\

\midrule

\multirow{4}{*}{{SAM 2.1}}
& {\textit{Tracker only}}
& {\textbf{95.6}} & {23.8} & {38.1}
& {84.4} & {21.0} & {33.6} \\

& {TubeletGraph \cite{sun2026tracking}}
& {\underline{95.1}} & {42.5} & {\underline{58.8}}
& {86.4} & {38.7} & {53.4} \\

& {\methodName \textit{w/o Judge}}
& {91.8} & {\underline{55.8}} & {\textbf{69.4}}
& {\underline{89.1}} & {\underline{54.1}} & {\underline{67.4}} \\

& {\methodName (Ours)}
& {90.3} & {\textbf{56.4}} & {\textbf{69.4}}
& {\textbf{89.4}} & {\textbf{55.8}} & {\textbf{68.7}} \\

\midrule

\multirow{4}{*}{SAM 3}
& \textit{Tracker only}
& {84.4} & {21.0} & {33.6}
& {66.7} & {16.6} & {26.5} \\

& TubeletGraph \cite{sun2026tracking}
& {89.8} & {\underline{24.3}} & {38.3}
& {71.4} & {\underline{19.3}} & {30.4} \\

& \methodName \textit{w/o Judge}
& {\underline{95.3}} & {\textbf{55.8}} & {\underline{70.4}}
& {\underline{95.3}} & {\textbf{55.8}} & {\underline{70.4}} \\

& \methodName (Ours)
& {\textbf{96.2}} & {\textbf{55.8}} & {\textbf{70.6}}
& {\textbf{96.2}} & {\textbf{55.8}} & {\textbf{70.6}} \\

\bottomrule
\end{tabularx}
\end{table*}

%% file: appendix/extended_comparisons.tex
{
\section{Extended Comparisons}
\label{apdx:extended}
This section complements the state-of-the-art comparison of \cref{sec:main_results} with additional experiments that {detail the two additional baselines,} provide fully open and reproducible results on \datasetName, {report the full comparison on VOST-TAS,} and compare with other trackers on VOST.

\input{appendix/baselines}

{
\subsection{Full Comparison on VOST-TAS}
\label{apdx:vost_tas_full}

\input{tables/vost_tas_summary}

\cref{tab:vost-tas_main} extends \cref{tab:vost_tas_hack} with \methodName, vanilla SAM, and the Agentic baseline on VOST-TAS~\cite{sun2026tracking}.
Interpreting performance on VOST-TAS requires understanding the critical structural flaws of its evaluation protocol:
first, VOST-TAS annotates only the resulting parts at the end frame of each transformation without modeling the state graph;
second, its primary metric $\mathcal{H}_{ST}$ is strictly recall-based and verifies spatial overlap only at the final frame of the ground-truth window, completely blind to which input objects participated in the event.
As demonstrated in \cref{tab:vost_tas_hack}, this flawed formulation heavily rewards blind over-prediction: the AlwaysPredict baseline achieves the highest $\mathcal{H}_{ST}$ (13.9 and 15.7) simply by firing spurious disassemblies at every mask split, despite a disastrous temporal precision of 6.2--9.0 $\mathcal{T}^{\text{TAS}}_P$.

This metric distortion explains the apparent disparity between TubeletGraph and \methodName under SAM~2.1.
VOST was specifically collected to induce catastrophic tracker failures through extreme blur and heavy object deformation~\cite{tokmakov2023breaking}.
Because TubeletGraph detects transformations by interpreting track loss as a structural split, it is artificially rewarded on VOST when a weaker tracker (SAM~2.1) frequently drops tracks, allowing its reactive tubelet search to trigger and accumulate unpenalized recall credit on $\mathcal{H}_{ST}$.
In contrast, \methodName evaluates geometry proactively and refuses to predict transformations unless coherent disjoint components and mask priors support them.
On severe visual artifacts where SAM~2.1 collapses, \methodName conservatively suppresses false positives, yielding a low $\mathcal{H}_{ST}$ (2.2) solely because the metric fails to penalize TubeletGraph's hallucinated proposals.

Once the base tracker provides reliable segmentation masks (SAM~3), TubeletGraph's failure-triggered assumption collapses: its performance drops by half (13.0 to 6.5 $\mathcal{H}_{ST}$), whereas \methodName directly translates the improved tracking quality into balanced precision and recall.
With SAM~3, \methodName is the only method exceeding 20 in both $\mathcal{T}^{\text{TAS}}_P$ and $\mathcal{T}^{\text{TAS}}_R$, while achieving the best tracking quality with a $+4.0$ $\mathcal{J}_{tr}$ improvement over TubeletGraph (40.7 vs 36.7).
This confirms that the lower $\mathcal{H}_{ST}$ of \methodName on SAM~2.1 is an artifact of a metric that rewards tracking failures, whereas VISTA provides the rigorous, object-grounded evaluation necessary to assess physical transformation tracking.
}

{
\subsection{TubeletGraph Failure Mode with SAM 3}
\label{apdx:tg_sam3_failure}

\input{figures/tg_sam3_failure}

Upgrading TubeletGraph's tracker from SAM 2.1 to SAM 3 reduces the $\mathcal{T}_{F_1}$ from 12.5 to 3.3 on \datasetName (\cref{tab:tuvos_main}) and $\mathcal{H}_{ST}$ from 13.0 to 6.5 on VOST-TAS{ (\cref{tab:vost-tas_main})}. \cref{fig:tg_sam3_failure} illustrates the cause of this drop on two sequences: wii\_remote\_disassembly from VISTA and 3161\_peel\_banana\_1 from VOST-TAS. With SAM 2.1, the tracked mask of the Wii remote rubber case is lost after the user extracts it, which triggers the tubelet search and leads TubeletGraph to detect the transformation. With SAM 3, the rubber case is tracked robustly throughout the video, the track loss occurs later on in the video and no transformation is ever triggered due to the proximity constraint not being satisfied. Similarly, when the battery cover is removed, SAM 3 keeps tracking it, so TubeletGraph cannot trigger any transformation event by design. As a result, the state graph collapses to a single node, producing a graph with no transformations. The banana example exhibits the same behavior, with all three peeling actions missed when using SAM 3. This explains why TubeletGraph cannot benefit from the stronger SAM 3, whereas \methodName translates the improved tracking quality into more accurate state graphs.
}

\subsection{Fully Open-weight Pipeline}
\label{apdx:open_pipeline}

\input{tables/open_pipeline}

In our default configuration, \methodName already employs an open-weight MLLM Judge (Qwen 3.5 35B), while the MLLM Labeler relies on the closed-source GPT-5.4. To provide a fully accessible and reproducible pipeline, we replace the MLLM Labeler with Qwen 3.5 35B, such that all the components of \methodName, as well as the LLM evaluator (Gemma 4 31B), are open-weight models.

As shown in \cref{tab:open_pipeline}, the fully open-weight configuration remains highly competitive. Compared to the default configuration in \cref{tab:tuvos_main}, replacing GPT-5.4 with Qwen 3.5 35B mainly affects action naming ($\mathcal{S}_A$), leading to a moderate decrease of the overall score ($25.5$ vs $27.1$ $\mathcal{H}_{F_1}$ with SAM 2.1, and $23.2$ vs $28.3$ with SAM 3). Nevertheless, \methodName still outperforms TubeletGraph by a wide margin, achieving $+22.9$ and $+23.2$ $\mathcal{H}_{F_1}$ points with SAM 2.1 and SAM 3, respectively.

\subsection{Tracking-only Comparison on VOST}
\label{apdx:vost_tracking}

\input{tables/vost_tracking}

To the best of our knowledge, no existing method natively addresses the \taskShortName task, and TubeletGraph is the only baseline able to detect and track transformations, albeit limited to disassemblies. While modern VOS models excel at propagating pixel-level masks, they cannot generate state graphs ($\mathcal{T}_{F_1}=0$). To nonetheless benchmark pure tracking capabilities, we compare \methodName against state-of-the-art VOS models on the VOST~\cite{tokmakov2023breaking} validation set in \cref{tab:vost_tracking}. SAM 3 remains a highly competitive base tracker, and \methodName with SAM 3 achieves the best tracking quality overall, improving over vanilla SAM 3 by $+1.5$ $\mathcal{J}$ and $+3.4$ $\mathcal{J}_{tr}$ points.
}

%% file: appendix/baselines.tex
{
\subsection{Additional Baselines}
\label{apdx:baselines}

This {subsection} provides {further details on the} alternative fair baselines {compared in \cref{sec:main_results}}.

\paragraph{AlwaysPredict {Baseline}.} This baseline assesses the feasibility of detecting object transformations without the \methodName specific modules. This pipeline keeps the tracking and component extraction and removes the point sampling, state generation, evaluation, selection, and MLLM Judge. When the number of components is greater than 1 a disassembly is predicted, and when two objects' masks overlap with an IoU above $0.5$ an assembly is predicted. {\cref{tab:tuvos_main}} shows that the AlwaysPredict baseline performs poorly, with {gaps} of {$-27.6$ and $-24.9$} $\mathcal{T}_{F_1}$ points compared with \methodName \textit{w/o Judge}, which likewise uses no MLLM. {Its precision drops to 6.6 and 8.2 $\mathcal{T}_{p}$ against 25.3 and 17.9 $\mathcal{T}_{r}$, indicating that most of the predicted transformations are spurious, mirroring the over-prediction observed on VOST-TAS (\cref{tab:vost_tas_hack}).} This demonstrates that tracker cues alone are insufficient to detect transformations, and that state generation and evaluation are required.

\paragraph{Agentic {Baseline}.} This baseline tests the MLLM Judge (Qwen~3.5~35B) under a different paradigm. Differently from the \methodName setting, we provide a full agentic harness with the following tools:

\begin{itemize}
    \item \textit{video and image analysis}: this includes tools like view of multiple frames, zooming and segmentation mask toggle.
    \item \textit{SAM 3 prompting}: this includes all the modalities to prompt SAM 3 (point, box, text) and the use of the tracker.
    \item \textit{transformations recording}: this includes tools to record transformations by providing object IDs with the corresponding semantic labels.
\end{itemize}

Notably, the harness operates offline: unlike \methodName, which processes the video in a streaming fashion, the agent can access any frame at any time, which gives it an advantage. The number of steps and tool calls is unlimited, so the agent runs until it decides to stop. \cref{fig:agent_prompt} presents the initial prompt provided to the agent.
Despite this advantage, {\cref{tab:tuvos_main}} shows {large gaps} of {$-34.8$ and $-32.2$} $\mathcal{T}_{F_1}$ points compared with \methodName, which uses the same MLLM, while requiring {$2.0\times$ and} $2.4\times$ the runtime of \methodName on the whole benchmark. {Unlike the AlwaysPredict baseline, the agent records only a few transformations, reaching 15.8 and 19.7 $\mathcal{T}_{p}$ but only 7.6 and 7.5 $\mathcal{T}_{r}$.} This demonstrates that giving the MLLM direct control over the tracker does not replace the structured state proposals of \methodName, even with access to the full video.

\begin{figure*}[p]

\centering
\begin{userprompt}
You are a video analysis agent. You control a video object tracker (SAM~3) through tools, and you can watch the video through tools. Your job is to produce a complete, accurate record of the physical objects in a video and of the transformations they undergo.

\paragraph{The task}
The video shows physical objects being taken apart or put together. Think of an \textbf{object} as anything that would move as a single unit if left alone: if you picked it up, everything that is part of it would come along.

A \textbf{transformation} happens when this changes:
\begin{itemize}
  \item \textbf{DISASSEMBLY}: one object becomes two or more objects.
  \item \textbf{ASSEMBLY}: two or more objects become one object.
\end{itemize}

Any change in what forms a unit counts, however small the part involved. Contact that does not outlast the hand (holding, touching, occlusion) does not change what forms a unit.

You must produce:
\begin{enumerate}
  \item \textbf{A track for every object} relevant to the task. A track is a segmentation mask on every frame where the object exists. A track is an empty mask when the object is fully occluded, and has no mask at all outside the object's lifetime.
  \item \textbf{Every transformation}, recorded at the frame where it happens (the first frame where the new state is visible). Each transformation lists:
  \begin{itemize}
    \item the objects before it (the ``before'' objects);
    \item the resulting objects;
    \item the action.
  \end{itemize}
  The ``before'' tracks end at that frame, and each resulting object starts a NEW track at that frame, even one that looks almost unchanged. Assembly/disassembly results are never continuations of an old track.

  Together, the tracks and transformations form a graph of how objects change over time. Results of a transformation are objects like any other: they can take part in later transformations (be split again, or assembled with other objects), so keep following every object until the end of the video.
  \item \textbf{Concise names} for every object (e.g.\ ``pen cap'', ``pen body'', ``assembled pen'').
  \item \textbf{The action of each transformation, described with VERBS only.} You may use several verbs (e.g.\ ``unscrew, pull apart'' or ``snap, press'').
\end{enumerate}
\end{userprompt}
\caption{MLLM Agent initial prompt (part 1 of 2).}
\label{fig:agent_prompt}
\end{figure*}

\begin{figure*}[p]
\ContinuedFloat

\centering
\begin{userprompt}
\paragraph{What you are given}
\begin{itemize}
  \item Some objects are \textbf{provided by the task} as initial masks: those visible at the start, and any object that later enters the scene and is not the result of a transformation. They are added to the tracker automatically when tracking reaches their first frame, and you are notified. Give each of them a concise name with \texttt{rename\_object}.
  \item Do NOT create tracks for objects that are not involved in the task (hands, tools that are not transformed, background, etc.). Only add new tracks yourself for the results of transformations (via \texttt{apply\_transformation}). Use \texttt{add\_object} only for an object you are sure is a task object that is missing.
\end{itemize}

\paragraph{How you are evaluated}
\begin{itemize}
  \item Mask quality of all tracks over time, especially at the end of the video.
  \item Correct detection of transformations: right type, right frame (timing matters), and right before/after objects. The result masks must match the real parts right after the event.
  \item Semantic accuracy of object names and action verbs.
  \item Missed and spurious transformations are penalised equally.
\end{itemize}

\paragraph{How the tools work}
\begin{itemize}
  \item The tracker has a \textbf{frontier}: the last frame that has been tracked. \texttt{propagate} moves it forward, and every edit happens at a frame $\leq$ frontier. Editing a frame behind the frontier automatically re-tracks the affected objects up to the frontier. Other tracks are not touched.
  \item Masks are created in two steps:
  \begin{enumerate}
    \item \texttt{segment} proposes candidates (\texttt{cand:<k>}) from points, a box or a text phrase. Nothing is committed yet.
    \item You commit a candidate with \texttt{add\_object}, \texttt{refine\_object}, \texttt{reactivate\_object} or \texttt{apply\_transformation}.
  \end{enumerate}
  Always check the preview before committing. You can also reference an existing track's mask with \texttt{obj:<id>@<frame>}, and combine masks with \texttt{combine\_masks}.
  \item Coordinates are relative to the FULL frame, normalised to 0--1000 on both axes ($(0,0)$ = top-left, $(1000,1000)$ = bottom-right).
  \item Video access mode: \textbf{offline}. You may look at any frame of the video at any time.
  \item Images are expensive. Use \texttt{view\_frames} with \texttt{layout="sheet"} to skim many frames at once, then look at single frames (\texttt{layout="separate"}) or \texttt{zoom} only where you need detail. Old images are removed from your context automatically, so write down important observations with \texttt{write\_notes}.
  \item Each tool result ends with a \texttt{[STATE]} block. It is the authoritative summary of frontier, tracks and transformations.
\end{itemize}

Act autonomously. Do not ask the user questions. Always respond with tool calls until you call \texttt{finish}.
\end{userprompt}
\caption{MLLM Agent initial prompt (part 2 of 2). Each episode then receives the video metadata and the provided masks at frame 0.}
\end{figure*}
}

%% file: tables/vost_tas_summary.tex
\begin{table*}[t]
\centering
\footnotesize

\caption{\textbf{State-of-the-art comparison on VOST-TAS.} We evaluate pure tracking and transformation recall using two foundational trackers. Best results per tracker are in \textbf{bold}, and second-best are \underline{underlined}.}
\label{tab:vost-tas_main}

\setlength{\tabcolsep}{4pt}
\renewcommand{\arraystretch}{1.05}

\begin{tabularx}{\linewidth}{
l
l
>{\centering\arraybackslash}X
>{\centering\arraybackslash}X
>{\centering\arraybackslash}X
>{\centering\arraybackslash}X
>{\centering\arraybackslash}X
>{\centering\arraybackslash}X
>{\centering\arraybackslash}X
>{\centering\arraybackslash}X
>{\centering\arraybackslash}X
}
\toprule

\multirow{2}{*}{Tracker} &
\multirow{2}{*}{Method} &
\multicolumn{2}{c}{\textsc{Tracking}} &
\multicolumn{2}{c}{\textsc{Sem. Sim.}} &
\multicolumn{2}{c}{\textsc{Temp. Loc.}} &
\multicolumn{2}{c}{\textsc{Overall}} &
\multirow{2}{*}{FPS $\uparrow$} \\

\cmidrule(lr){3-4}
\cmidrule(lr){5-6}
\cmidrule(lr){7-8}
\cmidrule(lr){9-10}

&
& $\mathcal{J}\uparrow$
& $\mathcal{J}_{tr}\uparrow$
& $\mathcal{S}_A\uparrow$
& $\mathcal{S}_O\uparrow$
& {$\mathcal{T}^{\text{TAS}}_P\uparrow$}
& {$\mathcal{T}^{\text{TAS}}_R\uparrow$}
& $\mathcal{H}_{ST}\uparrow$
& $\mathcal{H}\uparrow$
& \\

\midrule

\multirow{6}{*}{SAM 2.1}
& \blname{Vanilla SAM \cite{ravi2025sam}}
& \blname{48.4} & \blname{32.5}
& \blname{--} & \blname{--}
& \blname{--} & \blname{--}
& \blname{--} & \blname{--}
& \blname{\textbf{4.0}} \\

& {\blname{AlwaysPredict}}
& {\blname{48.8}} & {\blname{31.8}}
& {\blname{--}} & {\blname{--}}
& {\blname{6.2}} & {\blname{\textbf{51.9}}}
& {\blname{\textbf{13.9}}} & {\blname{--}}
& {\blname{2.0}} \\

& {\blname{Agentic}}
& {\blname{41.2}} & {\blname{25.9}}
& {\blname{\textbf{100.0}}} & {\blname{66.7}}
& {\blname{\textbf{45.0}}} & {\blname{8.3}}
& {\blname{0.0}} & {\blname{0.0}}
& {\blname{0.3}} \\

& TubeletGraph \cite{sun2026tracking}
& \underline{49.2} & \textbf{36.0}
& {63.6} & {88.2}
& {\underline{38.4}} & {\underline{23.1}}
& {\underline{13.0}} & \underline{{7.4}}
& 0.2 \\

&
\cellcolor{gray!10}\methodName \textit{w/o Judge}
& \cellcolor{gray!10}49.0
& \cellcolor{gray!10}33.6
& \cellcolor{gray!10}{{71.9}}
& \cellcolor{gray!10}\underline{{88.5}}
& \cellcolor{gray!10}14.1
& \cellcolor{gray!10}{13.3}
& \cellcolor{gray!10}{2.5}
& \cellcolor{gray!10}\textbf{{2.5}}
& \cellcolor{gray!10}\underline{3.1} \\

&
\cellcolor{gray!10}\methodName (Ours)
& \cellcolor{gray!10}\textbf{49.3}
& \cellcolor{gray!10}\underline{34.1}
& \cellcolor{gray!10}{\underline{{87.1}}}
& \cellcolor{gray!10}\textbf{{94.4}}
& \cellcolor{gray!10}{23.7}
& \cellcolor{gray!10}9.9
& \cellcolor{gray!10}2.2
& \cellcolor{gray!10}\underline{{1.9}}
& \cellcolor{gray!10}1.6 \\

\midrule

\multirow{6}{*}{SAM 3}
& \blname{Vanilla SAM \cite{carion2026sam}}
& \blname{53.3} & \blname{37.3}
& \blname{--} & \blname{--}
& \blname{--} & \blname{--}
& \blname{--} & \blname{--}
& \blname{\textbf{4.8}} \\

& {\blname{AlwaysPredict}}
& {\blname{52.5}} & {\blname{38.0}}
& {\blname{--}} & {\blname{--}}
& {\blname{9.0}} & {\blname{\textbf{47.2}}}
& {\blname{\textbf{15.7}}} & {\blname{--}}
& {\blname{2.8}} \\

& {\blname{Agentic}}
& {\blname{50.1}} & {\blname{31.6}}
& {\blname{\textbf{100.0}}} & {\blname{83.3}}
& {\blname{\textbf{29.6}}} & {\blname{7.4}}
& {\blname{0.9}} & {\blname{0.9}}
& {\blname{0.4}} \\

& TubeletGraph \cite{sun2026tracking}
& 52.3 & 36.7
& {{64.0}} & \textbf{{89.8}}
& {\underline{29.0}} & 10.2
& 6.5 & \underline{{1.9}}
& 0.3 \\

&
\cellcolor{gray!10}\methodName \textit{w/o Judge}
& \cellcolor{gray!10}\underline{53.6}
& \cellcolor{gray!10}\underline{38.2}
& \cellcolor{gray!10}{61.9}
& \cellcolor{gray!10}\underline{{89.1}}
& \cellcolor{gray!10}15.1
& \cellcolor{gray!10}{21.0}
& \cellcolor{gray!10}{8.3}
& \cellcolor{gray!10}{4.9}
& \cellcolor{gray!10}\underline{3.6} \\

&
\cellcolor{gray!10}\methodName (Ours)
& \cellcolor{gray!10}\textbf{54.8}
& \cellcolor{gray!10}\textbf{40.7}
& \cellcolor{gray!10}{\underline{{83.4}}}
& \cellcolor{gray!10}{88.5}
& \cellcolor{gray!10}{27.4}
& \cellcolor{gray!10}{\underline{24.1}}
& \cellcolor{gray!10}{\underline{12.3}}
& \cellcolor{gray!10}\textbf{{9.6}}
& \cellcolor{gray!10}1.6 \\

\bottomrule
\end{tabularx}
\end{table*}

%% file: figures/tg_sam3_failure.tex
\begin{figure*}[p]

    \centering
    \includegraphics[width=\textwidth]{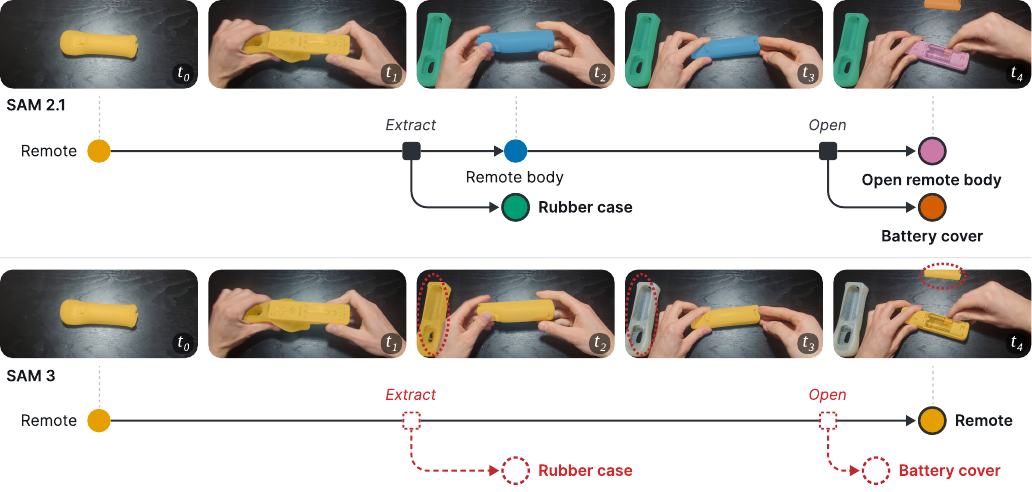}\\[6pt]
    \includegraphics[width=\textwidth]{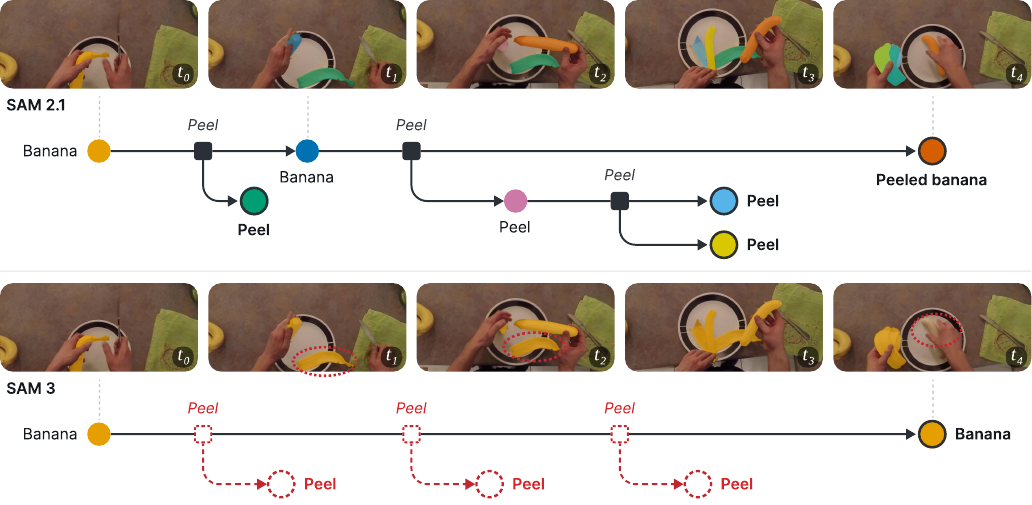}
    \caption{\textbf{TubeletGraph failure mode with SAM 3.} State graphs predicted by TubeletGraph with SAM 2.1 and SAM 3 as tracker on two sequences: \textbf{(top)} wii\_remote\_disassembly from VISTA, \textbf{(bottom)} 3161\_peel\_banana\_1 from VOST-TAS. Dashed red circles and boxes denote missed objects and transformations. With SAM 3, the tracks are lost later in the video, invalidating TubeletGraph's assumptions on proximity and on lost tracking. This directly translates to missed transformation detection.}
    \label{fig:tg_sam3_failure}
\end{figure*}

%% file: tables/open_pipeline.tex
\begin{table*}[t]

\centering
\footnotesize

\caption{\textbf{Fully open-weight pipeline on \datasetName.} \methodName uses Qwen 3.5 35B instead of GPT-5.4 as MLLM Labeler. Tracking ($\mathcal{J}$, $\mathcal{J}_{tr}$) and transformation tracking ($\mathcal{T}_{F_1}$) do not depend on the labeler and are identical to \cref{tab:tuvos_main}. Values are reported as mean $\pm$ standard deviation across three seeds. Best results per tracker are in \textbf{bold}.}
\label{tab:open_pipeline}

\setlength{\tabcolsep}{4pt}
\renewcommand{\arraystretch}{1.05}

\begin{tabularx}{\linewidth}{
l
l
>{\centering\arraybackslash}X
>{\centering\arraybackslash}X
>{\centering\arraybackslash}X
>{\centering\arraybackslash}X
>{\centering\arraybackslash}X
}
\toprule

\multirow{2}{*}{Tracker} &
\multirow{2}{*}{Method} &
\multicolumn{2}{c}{\textsc{Tracking}} &
\multicolumn{3}{c}{\textsc{State Graph}} \\

\cmidrule(lr){3-4}
\cmidrule(lr){5-7}

&
& $\mathcal{J}\uparrow$
& $\mathcal{J}_{tr}\uparrow$
& $\mathcal{S}\uparrow$
& $\mathcal{T}_{F_1}\uparrow$
& $\mathcal{H}_{F_1}\uparrow$ \\

\midrule

\multirow{2}{*}{SAM 2.1}
& \methodName \textit{w/o Judge}
& 90.4 $\pm$ 0.6 & 87.2 $\pm$ 0.9
& 73.8 $\pm$ 3.1 & 36.7 $\pm$ 3.3 & 19.2 $\pm$ 4.6 \\

& \methodName (Ours)
& \textbf{90.7 $\pm$ 0.3} & \textbf{88.6 $\pm$ 0.2}
& \textbf{74.8 $\pm$ 1.5} & \textbf{44.7 $\pm$ 1.6} & \textbf{25.5 $\pm$ 2.2} \\

\midrule

\multirow{2}{*}{SAM 3}
& \methodName \textit{w/o Judge}
& \textbf{91.5 $\pm$ 1.0} & \textbf{88.0 $\pm$ 1.4}
& 71.8 $\pm$ 2.8 & 33.4 $\pm$ 1.8 & 17.6 $\pm$ 0.8 \\

& \methodName (Ours)
& 91.4 $\pm$ 0.2 & 87.8 $\pm$ 0.5
& \textbf{74.0 $\pm$ 1.2} & \textbf{42.5 $\pm$ 1.5} & \textbf{23.2 $\pm$ 0.6} \\

\bottomrule
\end{tabularx}
\end{table*}

%% file: tables/vost_tracking.tex
\begin{table}[t]

\centering
\footnotesize
\caption{\textbf{Tracking-only comparison on VOST.} We compare \methodName against state-of-the-art VOS models on the VOST~\cite{tokmakov2023breaking} validation set. Only \methodName is able to detect both assembly and disassembly transformations, while TubeletGraph detects only disassemblies. Best results are in \textbf{bold}.}
\label{tab:vost_tracking}

\begin{tabular*}{\linewidth}{@{\hspace{6pt}\extracolsep{\fill}}lcc@{\hspace{6pt}}}
\toprule
Method & $\mathcal{J}\uparrow$ & $\mathcal{J}_{tr}\uparrow$ \\
\midrule
Cutie \cite{cheng2024putting} & 41.1 & 25.5 \\
ReVOS \cite{chen2025m3vos} & 41.0 & 25.3 \\
SAM 2 \cite{ravi2025sam} & 46.1 & 29.4 \\
SAM 2.1 \cite{ravi2025sam} & 48.4 & 32.5 \\
SAM2Long \cite{ding2025sam2long} & 46.4 & 29.1 \\
DAM4SAM \cite{videnovic2025distractor} & 48.8 & 33.6 \\
SAMURAI \cite{yang2024samurai} & 49.8 & 34.0 \\
TubeletGraph (SAM 2.1) \cite{sun2026tracking} & 49.2 & 36.0 \\
SAM 3 \cite{carion2026sam} & 53.3 & 37.3 \\
TubeletGraph (SAM 3) \cite{sun2026tracking} & 52.3 & 36.7 \\
\midrule
\methodName \textit{w/o Judge} (SAM 2.1) & 49.0 & 33.6 \\
\methodName (SAM 2.1) & 49.3 & 34.1 \\
\methodName \textit{w/o Judge} (SAM 3) & 53.6 & 38.2 \\
\methodName (SAM 3) & \textbf{54.8} & \textbf{40.7} \\
\bottomrule
\end{tabular*}
\end{table}

%% file: appendix/additional_qualitative.tex
\section{Additional Qualitative Results, Failure Cases and Limitations}
\label{apdx:qualitatives}

\input{figures/qualitatives_supp}
\input{figures/qualitatives_failures}

{This section complements the qualitative results of \cref{sec:main_results}. First, we show that \methodName is not limited to rigid objects with an example of a non-rigid transformation from VOST-TAS. Then, we present additional qualitative results on \datasetName and analyze the failure cases of \methodName. Finally, we discuss the current limitations of our approach.}

{\paragraph{Non-rigid Transformations.}
\label{apdx:nonrigid}
While \datasetName is designed around assemblies and disassemblies of rigid objects and does not specifically target non-rigid ones, neither the \taskShortName formulation nor \methodName is limited to rigid transformations. Since state proposals are generated and evaluated directly from the tracked masks, no assumption is made on the shape of the involved objects. \cref{fig:nonrigid_example} shows an example from VOST-TAS (sequence \texttt{3511\_unscrew\_jar\_1}) with SAM 3 as base tracker. \methodName correctly detects the cap being unscrewed from the jar and, crucially, the sauce squeezed onto the pasta as a new, highly deformable entity, whose mask matches the ground truth with an IoU of $0.92$ at the end of the transformation. Notably, the masks of the cap and of the sauce are temporarily lost during the video (\eg, while the pasta is being stirred), yet both objects retain their identity once they reappear and no spurious transformation is introduced. This confirms that \methodName does not require uninterrupted tracks, since structural changes are evaluated directly from the available masks rather than triggered by track loss. The example also exhibits the failure modes discussed {below. While} the sauce is still flowing out of the jar, the tracker repeatedly splits and re-merges the two masks, producing four spurious transformations, and the MLLM Labeler assigns generic names to some states (\eg, the jar is initially labeled as a bottle cap).}

\paragraph{{Additional Qualitative Results and Failure Cases.}}
\label{apdx:failures}
\cref{fig:quali_supp} presents additional qualitative results on \datasetName{, illustrating successful multi-step assemblies, while \cref{fig:quali_failures} shows two disassemblies} from the Assembly101~\cite{sener2022assembly101} subset. While \methodName handles these sequences robustly, these examples also highlight specific failure modes stemming from our direct reliance on upstream foundation models. 

First, the MLLM Labeler occasionally struggles with fine-grained semantics. Without external context (\eg, an instruction manual), visually ambiguous components are often assigned generic labels, such as identifying isolated IKEA drawer parts simply as ``blocks''{ (\cref{fig:quali_supp}, top row)}. 

Second, segmentation and tracking errors propagate directly into the state graph. As seen in {\cref{fig:quali_failures}}, the base tracker occasionally drops small parts (\eg, screws or wheels) immediately following a disassembly. Furthermore, the {bottom row of \cref{fig:quali_failures}} illustrates a hallucinated transformation caused by a segmentation artifact: the tracker merges a screwdriver into the toy truck mask. Because \methodName relies on these masks to generate state proposals, this forces an unnecessary structural update in the graph, even though the MLLM Judge correctly recognizes them as separate entities.

{\paragraph{Limitations.}
\label{apdx:limitations}
As illustrated by the failure cases above, the accuracy of \methodName is bounded by the quality of the base tracker, the promptable mask decoder, and the reasoning capabilities of the MLLMs. A part dropped by the tracker or a segmentation artifact can lead to a missed or spurious transformation, and visually ambiguous parts can receive generic labels. Moreover, since \methodName operates in an online, streaming fashion, the state graph is built causally, and a transformation cannot be revised retroactively once later frames are observed. This property is shared by prior work, in fact, TubeletGraph~\cite{sun2026tracking} also processes videos in a streaming fashion. Rather than recovering lost tracks, it tracks all candidate regions of the scene as tubelets and decides whether to include them in the state graph after a transformation, without ever correcting the transformations already added to the predicted state graph. Crucially, in \methodName an error in the state graph does not propagate to the rest of the video. Although a wrong transformation remains in the graph, subsequent state proposals are evaluated directly from the available masks, allowing \methodName to return to a state consistent with the observed scene. As shown in \cref{fig:nonrigid_example}, objects whose masks are temporarily lost retain their identity once they reappear and, despite the four spurious transformations caused by the flowing sauce, the objects present at the end of the video are correctly identified. As shown in \cref{tab:last_frame}, \methodName identifies the objects present on the last frame with a precision of $93.4$, outperforming TubeletGraph by $+30.1$ $F_1$ points (Appendix~\ref{apdx:last_frame}). The remaining gap lies in recall ($56.7$), mostly due to small parts that are not picked up by the tracker during a transformation. Promising directions to address these limitations include an offline variant of \methodName that exploits future frames to revise past transformations, and the re-detection of parts missed during a transformation to close the recall gap.}

{Finally,} as discussed in \cref{sec:main_results}, {the direct dependence on foundation models} is also a fundamental strength. By continuously evaluating structural proposals directly from the raw tracker outputs, rather than relying on reactive and failure-triggered search heuristics, \methodName is uniquely positioned to scale synergistically. As future vision and language foundation models improve in robustness, our framework will naturally and automatically inherit those gains without requiring architectural changes.

%% file: figures/qualitatives_supp.tex
\begin{figure*}[p]
    \centering
    \includegraphics[width=\textwidth]{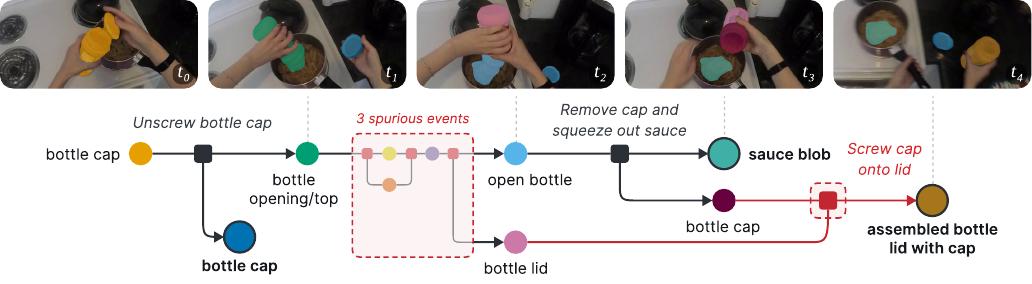}
    \caption{\textbf{Non-rigid transformation on VOST-TAS.} \methodName detects the sauce squeezed onto the pasta as a new, highly deformable entity and integrates it into the state graph. Node colors match the predicted masks in the frames, and bold nodes denote the objects present at the end of the video. Dashed red boxes denote spurious transformations.}
    \label{fig:nonrigid_example}

    \vspace{2em}
    \includegraphics[page=1, width=\textwidth,height=0.8\textheight,keepaspectratio]{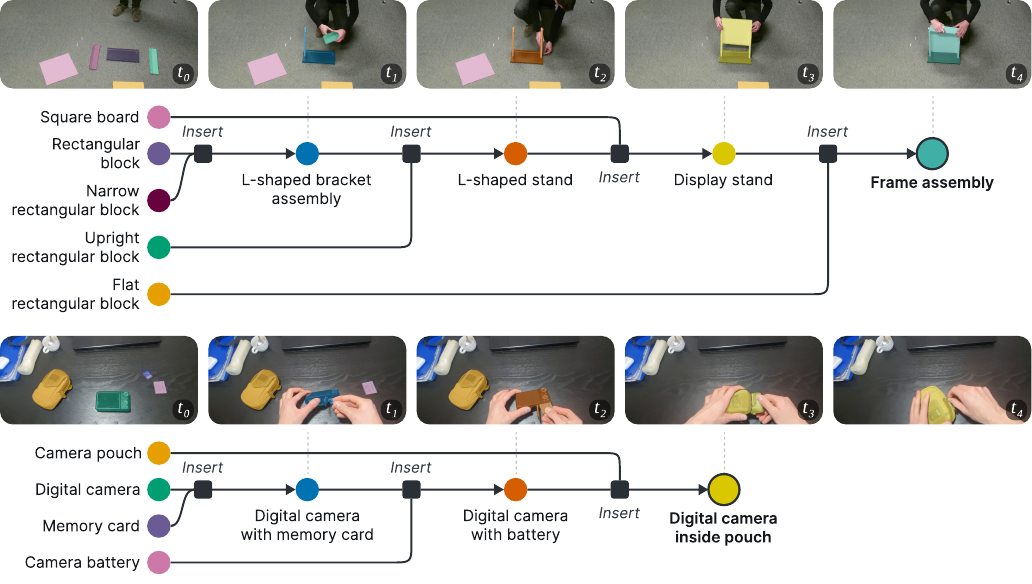}
    \caption{\textbf{Additional qualitative results of \methodName on the \datasetName benchmark.} We present {two assembly} sequences with their corresponding state-graph predictions. Five chronological video frames for each sequence are displayed above their respective state graph, illustrating how composite objects evolve over time. {The top row} details the multi-step construction of a drawer assembly from flat boards, while the bottom row {shows a digital camera that receives its memory card and battery before being inserted into its pouch. These examples demonstrate the capability of the framework to model multi-step assemblies and part-whole interactions.}}
    \label{fig:quali_supp}
\end{figure*}

%% file: figures/qualitatives_failures.tex
\begin{figure*}[p]
    \centering
    \includegraphics[page=1, width=\textwidth,height=0.8\textheight,keepaspectratio]{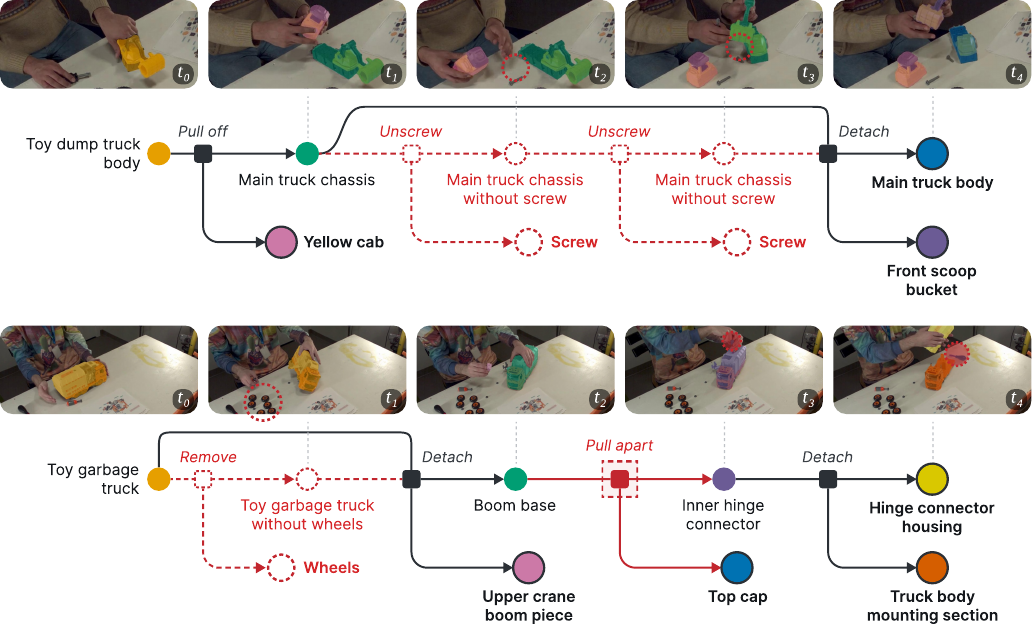}
    \caption{{\textbf{Failure cases of \methodName on the \datasetName benchmark.} We present two disassembly sequences from the Assembly101~\cite{sener2022assembly101} subset with their corresponding state-graph predictions. The top row shows the disassembly of a toy dump truck, while the bottom row features the complex disassembly of a toy garbage truck. Dashed red circles and boxes denote false negatives, \ie, objects and transformations present in the ground truth but undetected, such as the missed screws and wheels. A red action in a dashed red box denotes a false positive, \ie, a transformation incorrectly predicted by the pipeline, such as the hallucinated separation of the top cap.}}
    \label{fig:quali_failures}
\end{figure*}